\documentclass[letterpaper]{article} %
\usepackage[preprint]{aaai2027}  %
\usepackage[hyphens]{url}  %
\usepackage{graphicx} %
\usepackage{natbib}  %
\usepackage{caption} %
\usepackage{amsmath}
\usepackage{amsthm}

\usepackage{mathtools}
\usepackage{tabularx}
\usepackage{arydshln}
\usepackage{placeins}
\usepackage{dblfloatfix}
\usepackage{multirow}
\usepackage{amsfonts}
\usepackage{amssymb}

\usepackage{pifont}%
\newcommand{\cmark}{\ding{51}}%
\newcommand{\xmark}{\ding{54}}

\usepackage{algorithm}
\usepackage{algorithmic}
\usepackage{cleveref}

\usepackage{newfloat}
\usepackage{listings}
\DeclareCaptionStyle{ruled}{labelfont=normalfont,labelsep=colon,strut=off} %
\floatstyle{ruled}
\newfloat{listing}{tb}{lst}{}
\floatname{listing}{Listing}

\usepackage{booktabs}

\newcommand{\subfigurevspace}{\vspace{0mm}}

\RequirePackage{xspace}
\makeatletter
\DeclareRobustCommand\onedot{\futurelet\@let@token\@onedot}
\def\@onedot{\ifx\@let@token.\else.\null\fi\xspace}

\def\eg{\emph{e.g}\onedot} 

\def\ie{\emph{i.e}\onedot} 

\def\cf{\emph{cf}\onedot} 

\def\etc{\emph{etc}\onedot} 

\def\wrt{w.r.t\onedot}

\title{Specialist-Generalist Fusion with Outcome-Supervised Rationales \\ for Deepfake Detection}

\author {
	Benedikt Hopf\textsuperscript{\rm 1},
	Zongwei Wu\textsuperscript{\rm 1},
	Radu Timofte\textsuperscript{\rm 1}
}
\affiliations {
	\textsuperscript{\rm 1}Computer Vision Lab, CAIDAS, University of Würzburg, Germany\\
}

\begin{document}

\maketitle

\begin{abstract}

Generalizable deepfake detection requires complementary
forensic and semantic visual evidence. Specialist encoders
capture subtle manipulation traces but can overfit to
source-specific statistics, whereas MLLMs
provide broader visual-semantic representations but
may overlook fine forensic artifacts. We propose a two-stage
detector in which an MLLM directly fuses patch-level features
from a frozen forensic encoder with those from its native
vision encoder and produces the authenticity decision itself.
This specialist--generalist alignment provides the main
cross-domain performance gain.

We subsequently introduce outcome-supervised rationale
tuning. The model generates a free-form visual rationale
before its decision and is optimized using only the binary
authenticity label and a format constraint, without
task-specific rationale annotations. Rationale generation is
optional at inference, so the tuned model can still return a
direct binary score.

On DF40, specialist--generalist alignment improves average
cross-domain AUC from $89.85$ to $93.32\pm0.44$. Across six
paired runs, rationale tuning obtains $93.50\pm0.42$ and
improves five of six paired models; the mean difference is
small and not statistically conclusive. Results on SID-Set
and legacy benchmarks, together with fusion, output-order,
and continued-training controls, demonstrate the value of
complementary visual representations and show that
label-only rationale tuning can add an optional explanation
mode while approximately preserving direct detection
performance.

\end{abstract}
    
\section{Introduction}

\begin{figure}
    \centering
    \includegraphics[width=0.9\linewidth]{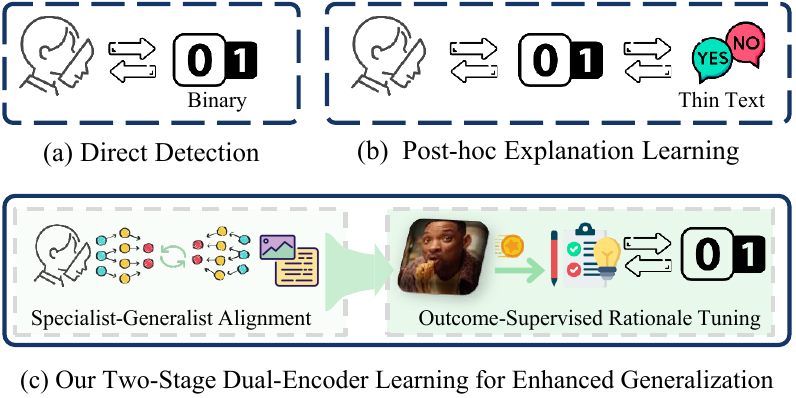}
    \caption{\textbf{Motivation:} Previous methods either (a) only use a specialized detector (with or without text) or (b) sequentially add a general MLLM after the detection step. Our approach jointly uses a generalist and specialist detector with an LLM as a fusion mechanism. In a second step, we use visual rationales as a form of auxiliary supervision to further refine performance and achieve optional explainability.}
    \label{fig:teaser}
\end{figure}

Generalizable deepfake detection requires complementary forms of visual evidence. Specialist forensic encoders are sensitive to subtle manipulation traces, but their predictions can depend on source- or generator-specific statistics. General-purpose multimodal encoders capture broader visual and semantic context, but are less reliable on fine forensic artifacts. Existing explainable systems~\cite{yu2025unlockingcapabilitieslargevisionlanguage, Sun2023TowardsGV, zhang2024commonsensereasoningdeepfake, guo2025rethinkingvisionlanguagemodelface} often use these components sequentially: a detector first produces a decision, after which an LLM explains it. Consequently, the LLM does not contribute to the detection decision itself.
Alternatively, recent work directly finetunes MLLMs for this task. This turns the vision encoder into the specialist, leading to a loss of broader semantic capabilities that could be less source-dependent.

In this work, we propose a solution (\cref{fig:teaser}), that fully utilizes the LLM without requiring it to become a full-capability deepfake detector by itself.
To achieve this, we leverage the intrinsic cross-modal alignment of MLLMs to combine features from a specialized detector and a general-purpose vision encoder. This dual-encoder design immediately provides frozen deepfake features, mitigating the overfitting in the general vision encoder. 
This integration alone significantly improves cross-dataset generalization and robustness to distribution shifts by performing effective fusion between the two input modalities. 

To further enhance generalization and simultaneously use the LLM for optional explanations, we formulate our objective through reinforcement learning (RL). 
Since deepfake detection is a binary task and most datasets provide only binary annotations, we use reinforcement learning to train free-form rationales based only on binary labels and a format restriction. Specifically, during training, the model generates a textual description highlighting potential manipulation artifacts and then predicts the image’s authenticity. The RL encourages descriptions that reliably lead to correct predictions while penalizing misleading or uninformative ones. 
Technically, we employ a variant of group-relative policy optimization~\cite{shao2024deepseekmathpushinglimitsmathematical} to sample multiple candidate explanations per input, which are ranked and rewarded based on correctness and format requirements. 

Different from previous CoT in mathematical reasoning, deepfake detection does not profit from intermediary results, but is rather limited by the need to find descriptive artifacts. Still, we show that meaningful rationales emerge under RL finetuning, despite not having a positive effect on training accuracy. Therefore, at inference time, producing descriptions is optional. Here we show a tradeoff between the two options: While generating rationales is advantageous for interpretability, direct binary classification has advantages in terms of speed and calibration. 

In summary, we propose a dual-encoder architecture that significantly outperforms the state-of-the-art in a binary training setting. Furthermore, we explore the use of training-time free-form rationales without task-specific rationale annotations to enhance interpretability and slightly improve performance, without requiring special dataset properties.

\paragraph{Contributions.}
Our contributions are threefold:

\begin{enumerate}
    \item \textbf{Direct specialist--native-vision fusion.}
    We introduce a deepfake detector that jointly processes
    patch-level representations from a frozen forensic encoder
    and the MLLM's native vision encoder. The authenticity
    decision is produced directly by the language model,
    allowing both visual streams to contribute to prediction.

    \item \textbf{Outcome-supervised optional rationales.}
    Starting from the binary-aligned detector, we train the
    model to generate a free-form visual rationale before its
    authenticity decision using only the binary label and a
    format constraint. This enables image-conditioned
    rationales without task-specific rationale annotations,
    while retaining direct one-token inference.

    \item \textbf{Cross-domain evaluation and controlled
    analysis.}
    On DF40, specialist--generalist alignment reaches
    $93.32\pm0.44$ average AUC compared with 89.85 for the
    specialist baseline. Across six paired runs, subsequent
    rationale tuning reaches $93.50\pm0.42$ and improves five
    of six pairs, although the average difference is not
    statistically conclusive. Fusion, continued-training,
    empty-rationale, and output-order controls characterize
    the contributions of the two training stages.
\end{enumerate}

\section{Related Work}

While deepfake detectors have progressed from simple general-purpose detectors~\cite{roessler2019faceforensicspp} to more complex ones like frequency-based~\cite{Liu2021SPSL, Luo2021GeneralizingFFHighFrequencyDetails, Hasanaath2024FSBI, Masi2020TwobranchRN, Qian2020F3Net}, spatial~\cite{Yan2023UCFUC, Yan2023LSDA, Cao2022RECCE, He2021BeyondTS, Guo2023ControllableGF, Zhuang2022UIAViTUI, yan2025orthogonalsubspacedecompositiongeneralizable, cui2025forensicsadapterunleashingclip}, pseudo-fake-based~\cite{Li2019FaceXRay, Zhao2020PCL, Shiohara2022SBI, nguyen2024laa, Larue2022SeeABLESD, hopf2025practicalmanipulationmodelrobust, nguyen2025vulnerabilityawarespatiotemporallearninggeneralizable} and video-based~\cite{kim2025spatialfrequencypixelwisetemporal, zheng2021ftcn, yan2024generalizingdeepfakevideodetection} they still generally only output a single binary label. Recently, the rise of multi-modal large language models has inspired several works~\cite{zhang2024commonsensereasoningdeepfake, yu2025unlockingcapabilitieslargevisionlanguage, guo2025rethinkingvisionlanguagemodelface, Sun2023TowardsGV} to employ language models as supervised deepfake detectors. This can potentially improve interpretability, but supervised training has the problem that ground-truth explanations are generally not available. 
Therefore, all these methods have their individual downsides, originating mainly from the fact that human annotations are costly and error-prone~\cite{Sun2023TowardsGV}, while hand-crafted heuristics are limited to a fixed number of categories determined by the authors. Furthermore, they treat text as an afterthought instead of fully incorporating it into the main model, missing any regularizing effects it might have.

Related work in general AI-generated content detection has also explored the use of language through annotated datasets~\cite{Huang2024SIDA, zhou2025aigiholmesexplainablegeneralizableaigenerated}. In particular some related works use GRPO~\cite{deepseekai2025deepseekr1incentivizingreasoningcapability, shao2024deepseekmathpushinglimitsmathematical} to obtain chain-of-thought~\cite{wei2023chainofthoughtpromptingelicitsreasoning} descriptions. Most methods~\cite{tan2025veritasgeneralizabledeepfakedetection, huang2025thinkfakereasoningmultimodallarge, huang2025sofakebenchmarkingexplainingsocial, jiang2026ivyfakeunifiedexplainableframework, Li_2026_OmniFake, shen2025authguardgeneralizabledeepfakedetection, nguyen2026prpoparagraphlevelpolicyoptimization, xu2026maremultimodalalignmentreinforcement, zhang2026cultivatingforensicreasoninggeneralizable} still require language annotations and are therefore not applicable to every dataset. While some of them can show that language supervision can improve generalization performance, we explore whether this also holds if no ground-truth annotations are available, and training can only use binary labels. In particular~\cite{huang2026realigngeneralizableimageforgery} distill the generalizable features learned from language into a separate non-LLM detector. We propose to train a model where reasoning is possible, but optional, without an additional distillation step. Key to our design is a novel special-general dual-encoder design that allows a generalist LLM to detect deepfakes without overfitting despite its large capacity and lack of textual supervision.
A few methods can -- similarly to our method -- be used without language annotations. However, \cite{Li2025RAIDXAR} is not built for partially fake images, ruling it out for deepfakes, which are, in many cases, partial face swaps. Furthermore, they rely on retrieval augmentation, which provides limited detection performance compared to our dual-encoder design. Finally, \cite{zhu2026evoguardextensibleagenticrlbased} build an agentic setup, dynamically calling upon multiple different detection tools to mimic a human user, moving more into the direction of orchestrating individual tools, rather than building a single detector.

\section{Method}

The fundamental challenge in deepfake detection is learning features that generalize well under distribution shifts, as in-dataset performance can generally easily achieve strong levels. Our central hypothesis is that a combination of general and specific features can limit overfitting by combining complementary modalities. We provide an MLLM with specialized features from a pretrained deepfake detector as well as features from the LLM's native vision encoder. We then obtain the decision from the output of the MLLM, rather than a separate classification head, to use the MLLM's pretrained knowledge to judge how to combine features. While this architecture already achieves the main improvement under binary pretraining alone, our experiments indicate that performing outcome-supervised rationale tuning can further improve the learnt representations, and simultaneously enable explainability on datasets with no language annotations.

\subsection{Continuous binary scores from a causal LM}

To be able to calculate all common metrics (\eg AUC), we need logits from a language model. Previous works' solution of using a separate classification head is not possible, as we want to fully involve the LLM for the feature fusion. Furthermore, binary answers cannot be used for AUC as well as any calibration metrics, and asking for continuous output (as a string) is complicated and error-prone. We therefore resort to the simple but effective solution of defining a target word (\textit{yes/no}) and its placement in the string and then accessing the full logits for the two words at that position. After a softmax, this allows for the full, continuous output.

\subsection{Stage 1: Specialist-Generalist Detection Alignment}
\begin{figure}
    \centering
    \includegraphics[width=0.9\linewidth]{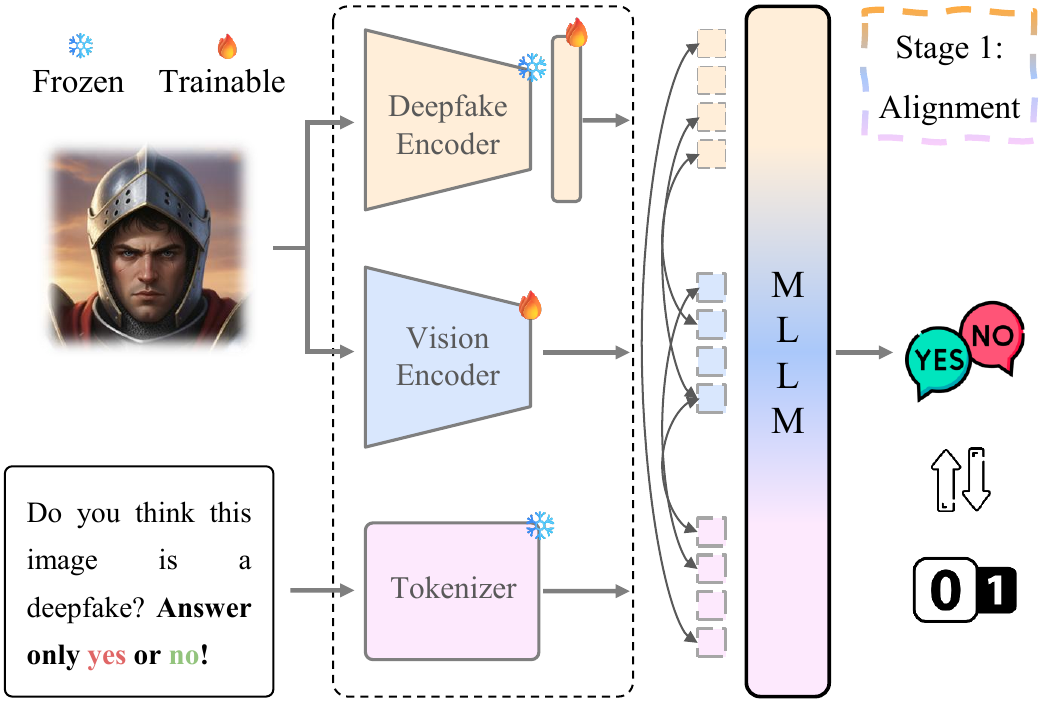}
    \caption{\textbf{First stage: Specialist-Generalist Detection Alignment.} Tokens from deepfake detector, vision encoder, and text are passed to the model, asking for a one-word answer. The model outputs a probability distribution over all words, which we can use for binary supervision and metrics. This alignment of the two complementary encoders results in the main performance improvement.
    }
    \label{fig:sft}
\end{figure}

Generalizable deepfake detection requires complementary forms of visual evidence. Specialist forensic encoders are sensitive to subtle manipulation traces, but their predictions can depend on source- or generator-specific statistics. General-purpose multimodal encoders capture broader visual and semantic context, but are less reliable on fine forensic artifacts. Existing explainable systems often use these components sequentially: a detector first produces a decision, after which a language model explains it. Consequently, the language model does not necessarily contribute to the detection decision itself. Alternatively, they finetune the generalist encoder, causing it to become a specialist, and therefore lose the second branch of visual evidence.

To enable the LLM to combine specific and general features, we use a dual-encoder design and obtain the output directly from the LLM. An overview of the architecture can be seen in \cref{fig:sft}. By not using a separate classification head, we involve the LLM in binary decisions.

We use two vision encoders: a pretrained deepfake detector $D$ as a specialized feature extractor and a vision-language model $L$ with vision-encoder $V$. We refer to the detector's last hidden state for some image $x$ as $D(x) \in \mathbb{R}^{n_D \times c_D}$, where $n_D$ and $c_D$ are the number of patches and channel dimension of the latent space of $D$. The same holds for $V(x) \in \mathbb{R}^{n_V \times c_V}$. The language model outputs logits $y = L(t_1, t_2, ..., t_n) \in \mathbb{R}^{n_L \times c_L}$ for input tokens $t \in \mathbb{R}^{n \times c_V}$ generated by $T$ from some input string $s$. %
Since in general $c_D \neq c_V$, we apply an affine transformation $f: \mathbb{R}^{c_D} \to \mathbb{R}^{c_V}$, aligning feature dimensions. This allows us to concatenate the deepfake, LLM-native vision, and language features. The entire forward pass can therefore be written as \begin{equation}
    y(s, x) = L(f(D(x)), V(x), T(s))
\end{equation}
We keep $D$ pretrained and frozen and only train $f$. Furthermore, all linear layers in $L$ and $V$ are finetuned using quantized low-rank adaptation~\cite{Hu2021LoRALA, dettmers2023qloraefficientfinetuningquantized} in order to keep memory requirements under control. Different from other finetuning works, our model immediately has specialist deepfake features. In combination with the LLM's native vision encoder, this creates an architectural prior that $V$ does not need to recreate the specificity of $D$.

\vspace{2mm}
\noindent \textbf{Supervision.} In this first step, we only use the LLM as a module for feature fusion, and therefore do not require full language output, but just a one-word answer. This allows direct, full supervision without language annotations. The prompt is always the same, asking for a one-word answer if the shown image is a deepfake. Here we can use SFT with standard cross-entropy loss, as we know the full answer (\textit{yes} or \textit{no}), since there is no description. 
This training step is comparably lightweight, as it requires only one target token rather than a multi-token rollout and requires only one generation per image. At the end of that step, the model has a unified internal representation of both modalities. Our analyses show that this incorporation of these complementary visual features significantly improves performance, without using the LLM's full language capabilities.

\begin{figure*}[tb]
    \centering
    \includegraphics[width=0.95\linewidth]{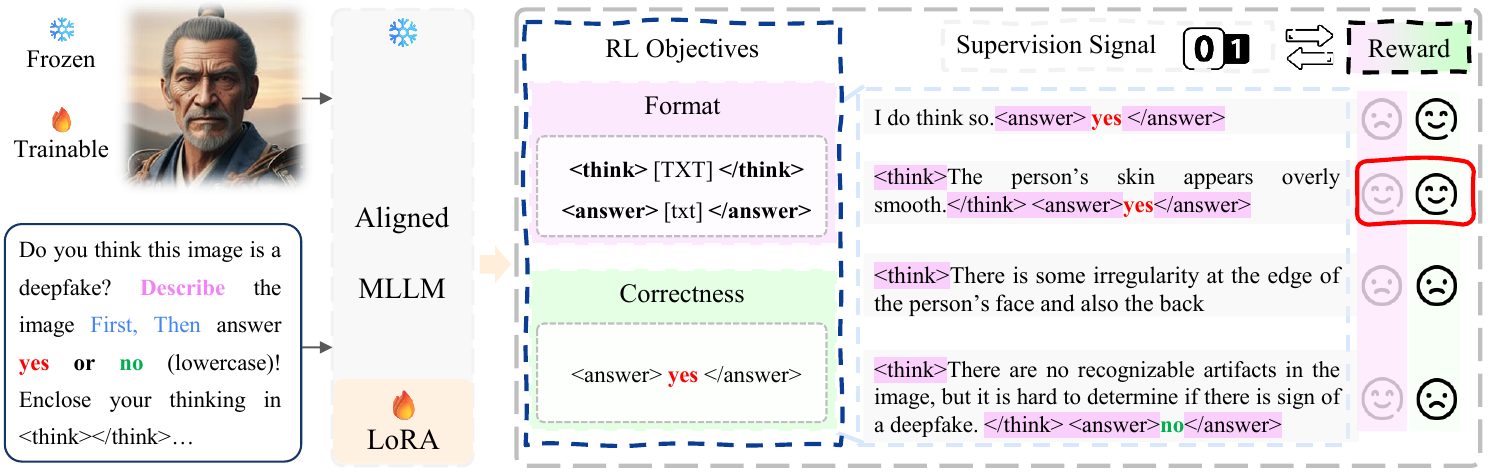}
    \caption{\textbf{Second stage: Outcome-Supervised Rationale Tuning.} We provide the model with the requested output structure, a question regarding the authenticity of the candidate image, and the image itself. We then sample multiple answers, judge them using our reward functions, and train using RL~\cite{shao2024deepseekmathpushinglimitsmathematical}. Negative advantages are discouraged, positive ones encouraged. 
    This enables generating free-form rationales at inference time for better interpretability of the model's decision. Furthermore, our experiments suggest a slight improvement in detection performance from the additional training signal.
    }
    \label{fig:grpo}
    \subfigurevspace
\end{figure*}

\subsection{Stage 2: Outcome-Supervised Rationale Tuning}

Once the two visual streams have been aligned for detection, we ask whether producing a visual rationale before the answer can serve as useful auxiliary supervision. Because rationale annotations are unavailable for most deepfake datasets, we optimize the generated sequence using the final authenticity label and a format constraint rather than target explanations. An overview of the RL pipeline is provided in \cref{fig:grpo}.
For the format reward $r^f_g$ of group item $g \in \mathcal{G}$ we simply define \begin{equation}
    r^f_g = \left\{ \begin{array}{ll}
        1 & \text{if $s_g$ matches $R$} \\
        0 & \text{else}
    \end{array} \right.
\end{equation} for sample string $s_g$ and regular expression $R$ as defined in \cref{eq:regex}.
\begin{equation}
\label{eq:regex} 
\begin{array}{l}
    \verb|<think>.*</think>| \\
    \verb|<answer>[a-z]+</answer>|
\end{array}
\end{equation}
The correctness reward $r^c_g$ for the predicted fake probability $p_g$ and correct label $l$ is given as \begin{equation}
    r^c_g = \left\{ \begin{array}{ll}
        1-|l-p_g| & \text{if $s_g$ ends with \texttt{</answer>}} \\
        -1 & \text{else}
    \end{array} \right.
\end{equation}
From the individual rewards, we calculate a combined final reward $r_g = r^f_g + \alpha \cdot r^c_g$, where $\alpha$ is a hyperparameter. 
We then use a slightly modified version of GRPO~\cite{deepseekai2025deepseekr1incentivizingreasoningcapability} for training (see suppl. for details).

We find that using $r^c$ before $r^f \approx 1$ causes instability and failure to follow the format. We therefore set $\alpha = 0.01 << 1$, such that $A \approx A^f$ unless $r^f_g = r^f_{g'} \ \forall g, g' \in \mathcal{G}$, in which case, normalization rescales $r^c$ to have the same magnitude, regardless of the choice of $\alpha$ (see supplementary for details).

This usage of language provides an auxiliary training output that empirically improves detection performance slightly. Our experiments show that this explain-then-classify approach is responsible for this improvement, rather than RL alone. We hypothesize that the activation of individual visual concepts during description reduces reliance on overly specific and non-generalizable artifacts. Furthermore, these rationales can also be used for enhanced interpretability. Their quality is discussed in the experiments section.

\subsection{Inference Modes}
\label{sec:inference_modes}

Given the two training stages, our model can operate in two modes for detection: One-word answer and explanation-based. Contrary to mathematical CoT reasoning, deepfake detection is not inherently a multi-step process, so we find reasoning at inference time to not be necessary and, in fact, detrimental for two reasons: 1. Computational demand: Reasoning requires multiple autoregressive decoding steps, while binary answers can be produced in one step. 2. Logit-quality: If the answer is conditional on an explanation, the final decision's probability is almost always pretty certain (\ie approximately 100\% or 0\%), losing the ability to provide meaningful intermediate answers. We therefore perform quantitative experiments in binary mode. In a traditional multi-step reasoning application, this would negate any advantage of reasoning, as it is not used at runtime, strengthening our proposal of using textual rationales only as an auxiliary training signal.

\begin{table*}[tb]
\centering
\begingroup
\setlength{\tabcolsep}{5pt}
    \centering

    \begin{tabular}{c||ccc|ccc|ccc|cccc} \toprule
         in \% & \multicolumn{3}{c|}{FS (cdf)} & \multicolumn{3}{c|}{FR (cdf)} & \multicolumn{3}{c|}{EFS (cdf)} & \multicolumn{4}{c}{average} \\ \hline
         Model & ACC$\uparrow$ & AUC$\uparrow$ & AP$\uparrow$ & ACC$\uparrow$ & AUC$\uparrow$ & AP$\uparrow$ & ACC$\uparrow$ & AUC$\uparrow$ & AP$\uparrow$ & ACC$\uparrow$ & AUC$\uparrow$ & AP$\uparrow$ & ECE$\downarrow$\\ \hline
         Xception & 77.86 & 76.23 & 88.91 & 79.70 & 83.05 & 94.00 & \underline{83.43} & 68.04 & 89.15 & \underline{80.33} & 75.78 & 90.68 & \underline{18.89} \\
         UIA-ViT & 76.21 & 87.57 & 95.24 & 78.13	& 89.22 & \underline{95.87} & 82.98 & 82.10 & 95.28 & 79.11 & 86.30 &  95.46 & 19.95 \\
         CLIP-L & \underline{76.44} & \textbf{94.23} & 95.36 & 78.55 & \underline{89.57} & 95.36 & 82.68 & \underline{85.76} & \underline{95.66} & 79.22 & \underline{89.85} & \underline{95.46} & 20.79 \\ 
         Effort & 76.40 & 90.78 & \underline{96.18} & \underline{78.61} & 81.76 & 93.47 & 82.49 & 78.85 & 94.26 & 79.17 & 83.80 & 94.64 & 20.85\\

         \midrule
         \textbf{Ours} & \textbf{85.81} & \underline{93.38} & \textbf{97.60} & \textbf{82.86} & \textbf{91.30} & \textbf{97.32} & \textbf{90.99} & \textbf{95.83} & \textbf{99.06} & \textbf{86.56} & \textbf{93.50} & \textbf{97.99} & \textbf{4.69}
    \end{tabular}
    \caption{\textbf{Comparison to the state-of-the-art on DF40.} Best values are highlighted in \textbf{bold}, second best values are \underline{underlined}. Particularly strong improvements are on EFS and accuracy. CLIP-L refers to the best model presented in~\cite{yan2024df40}}
    \label{tab:sota}
    \subfigurevspace
\endgroup
\end{table*}
\section{Experiments}

We provide both quantitative and qualitative results, ablations, and comparisons on various benchmark datasets.
\subsection{Experimental Settings}

\noindent\textbf{Datasets.} We perform our experiments in the framework of \textit{DeepfakeBench}~\cite{Yan2023DeepfakeBenchAC}, aligning with their evaluation protocol. We mainly use the dataset suite DF40~\cite{yan2024df40}, because it contains the three main fake types: face-swaps (FS), face-reenactment (FR), and entire-face synthesis (EFS), including very recent methods (40 in total). Furthermore, the benchmark is built on two different data sources, enabling cross-dataset testing for very flexible evaluations. We also provide results on SID-Set~\cite{Huang2024SIDA}, a fully-synthetic general-object deepfake detection dataset. This allows us to evaluate our model on a very modern dataset and compare to RAIDX~\cite{Li2025RAIDXAR}, which also creates rationales without textual supervision.

\vspace{2mm}
\noindent\textbf{Methods.} Due to DF40's recency, most methods have not been evaluated on this setup. We therefore choose methods, that can be trained on DF40, ranging from baselines to recent SOTAs: Xception~\cite{Chollet2016XceptionDL, roessler2019faceforensicspp}, UIA-ViT~\cite{Zhuang2022UIAViTUI}, CLIP-L (from DF40)~\cite{yan2024df40, radford2021clip}, and Effort~\cite{yan2025orthogonalsubspacedecompositiongeneralizable}. For SID-Set~\cite{Huang2024SIDA} we compare to LNP~\cite{Bi2023DetectingGI}, AntifakePrompt~\cite{Chang2023AntifakePromptPV}, SIDA~\cite{Huang2024SIDA} and RAIDX~\cite{Li2025RAIDXAR}. Legacy comparisons are in the supplementary.

\vspace{2mm} 
\noindent\textbf{Metrics.} Aligning with the common evaluation protocol~\cite{Yan2023DeepfakeBenchAC, yan2024df40, Zhuang2022UIAViTUI, yan2025orthogonalsubspacedecompositiongeneralizable, zhang2024commonsensereasoningdeepfake, yu2025unlockingcapabilitieslargevisionlanguage, guo2025rethinkingvisionlanguagemodelface, Sun2023TowardsGV}, we mainly report the area under the receiver-operating characteristic curve (AUC) and additionally accuracy, average precision, F1, and ECE~\cite{nguyen2024laa, roessler2019faceforensicspp, Li2025RAIDXAR}.

\vspace{2mm}
\noindent\textbf{Implementation Details.} We implement our model in the framework of~\cite{Yan2023DeepfakeBenchAC}, using their data preprocessing, metric calculation, \etc, allowing us to restrict improvements to the actual model architecture and training scheme, rather than spurious changes from different sources. %
We use Qwen2.5-VL-7B~\cite{bai2025qwen25vltechnicalreport} as our MLLM and CLIP-L from DF40~\cite{yan2024df40} as the deepfake detector.

\vspace{2mm}
\noindent\textbf{Computational Requirements.} 
As all language-output detectors rely on an LLM (\eg \cite{guo2025rethinkingvisionlanguagemodelface, yu2025unlockingcapabilitieslargevisionlanguage} also use a 7B model, and \cite{Sun2023TowardsGV} even uses a 13B model), training is slower than for binary detectors.
While repeated sampling during RL causes image throughput during training to be low, due to the alignment stage, fewer training steps are required in RL mode.
We train our model on Nvidia L40 GPUs (48 GB VRAM). For supervised training, we use a single GPU and a batch size of 4, with training taking less than 24h. For RL, we can only use a per-GPU batch size of 1. We therefore use 3 GPUs in parallel. Given the SFT pretraining, training is relatively fast, with most runs converging in less than 24h.

\subsection{Comparison to SOTA}
\label{sec:sota}

\noindent\textbf{Evaluation on DF40.} Here, we train on the full training split of DF40~\cite{yan2024df40}, which is based on real images from FF++~\cite{roessler2019faceforensicspp}. Evaluation is then performed cross-domain on images based on CDF~\cite{Li2019CelebDFv2}. This setting is particularly hard, as it requires cross-dataset generalization. We report averages of 6 runs of our full pipeline.
Our approach improves generalization compared to the state-of-the-art (\cf \cref{tab:sota}), beating the previous SOTA CLIP-L~\cite{yan2024df40} by $3.65\%$ AUC. We also see improvements on accuracy ($7.3\%$) and AP ($2.5\%$) on average over the different fake categories. The particularly large improvement on accuracy suggests that our method is naturally better calibrated than the state-of-the-art, which is confirmed by the expected calibration error (ECE) of $4.69\%$ compared to about $20\%$ for previous work. When looking at different types of fakes, it stands out that our method performs particularly well on fully synthetic images (+10\% AUC). We attribute this to the particularly strong risk of overfitting to low-level artifacts in these kinds of images compared to \eg face-swaps~\cite{Wang2019CNNGeneratedIA}. Our best run achieves an average AUC of $93.93\%$.

\vspace{2mm}
\noindent\textbf{Evaluation on SID-Set.} As an example of a modern state-of-the-art fully-synthetic general image dataset, we train and test on the fully-generated subset of SID-Set~\cite{Huang2024SIDA} (\cref{tab:sota-sid}), aligning with the evaluation protocol of RAIDX~\cite{Li2025RAIDXAR}. The comparison to RAIDX is particularly interesting, as it is the only other work we know of that can create language outputs without task-specific rationale annotations. While we both use RL, our dual-encoder design and binary pretraining prove more effective than RAIDX's retrieval augmentation. Our model outperforms all previous work, achieving $99\%+$ on both metrics and splits.
\begin{table}[tb]
    \small
    \centering

    \setlength{\tabcolsep}{1.9pt}
    \begin{tabular}{cc|cccc:cc} \toprule
    \multirow{2}{*}{\textbf{Model}} & \multirow{2}{*}{\textbf{Venue}} & \multicolumn{2}{c}{\textbf{Real}} & \multicolumn{2}{c}{\textbf{Fake}} & \multicolumn{2}{c}{\textbf{average}}\\
    & & ACC$\uparrow$ & F1$\uparrow$ & ACC$\uparrow$ & F1$\uparrow$ & ACC$\uparrow$ & F1$\uparrow$ \\ \midrule
    LNP & arXiv'23 & 71.2 & 83.2 & 91.8 & 95.7 & 81.5 & 89.5 \\
    AntifakePrompt & arXiv'23 & 64.8 & 78.6 & 93.8 & 96.8 & 79.3 & 87.7\\
    SIDA-13B & CVPR'25 & 96.7 & 97.3 & 98.7 & 99.3 & 97.7 & 98.3 \\
    RAIDX & MM'25 & \underline{98.5} & \underline{98.9} & \underline{99.4} & \underline{99.5} & \underline{99.0} & \underline{99.2} \\ \hline
    \textbf{Ours} & - & \textbf{99.4} & \textbf{99.7} & \textbf{99.6} & \textbf{99.8} & \textbf{99.5} & \textbf{99.7}\\
    \bottomrule
    \end{tabular}
    \caption{\textbf{Comparison on fully synthetic images.} Best values are highlighted in \textbf{bold}, second best values are \underline{underlined}; all values in \%. Our model outperforms previous methods. In particular, we outperform RAIDX~\cite{Li2025RAIDXAR}, due to our dual-encoder design and binary alignment-stage.}
    \label{tab:sota-sid}

    \subfigurevspace
\end{table}

\vspace{2mm}
\noindent\textbf{User Study.} We performed a user study, comparing some of our explanations to explanations from DD-VQA~\cite{zhang2024commonsensereasoningdeepfake} (human annotations). Users scored explanations on a scale from 1 (bad) to 5 (good) depending on whether the described artifacts exist in the image; details are provided in the supplementary. Our method achieves an average score of 3.57 with a standard deviation of 1.1, compared to DD-VQA with 3.83 and a standard deviation of 0.99, indicating that our method performs similarly to human annotations. 

\vspace{2mm}
\noindent\textbf{Text Quality.} We test for hallucinations by generating rationales on 40 images, which have human annotations from DD-VQA~\cite{zhang2024commonsensereasoningdeepfake}, and asking Google's Gemini to judge if there is any overlap between the explanations, indicating that both answers refer to the same artifact. We find a hallucination rate of only $7.5\%$, indicating that most of our model's answers are based on facts. This was expected because otherwise, the reasoning training could not conceivably improve detection performance. We evaluate artifact-level precision rather than exact wording or exhaustive localization: an explanation is counted as useful only when at least one claimed artifact is visibly supported by the image.

Additionally, we gave Gemini images along with the explanations and asked for a $0$--$5$ image-match score. Ours obtains $3.61\pm0.99$ vs. DD-VQA $3.79\pm0.78$. Gemini flagged two partial hallucinations among 20 explanations; both also contained another visible artifact, so we report them as partial errors rather than fully correct cases. 
On this 20-explanation claim-level check, Gemini marked 95.7\% of our artifact claims as image-supported, close to 96.4\% for DD-VQA. 

\vspace{2mm}
\noindent\textbf{Textual variety.}
We calculate self-BLEU to check for repetitiveness in our model's outputs. Here higher numbers indicate more repetitions/similarity between explanations. Our model achieves a self-BLEU value of $0.085$, which is considerably lower than the explanations from DD-VQA at $0.214$. This shows that our model generates diverse, individual rationales, rather than repetitive template sentences.

\subsection{Ablation and Discussion}

\noindent\textbf{Key Component Analysis.}
Our ablation study is shown in~\cref{tab:abation-sft}. 
Starting from the baseline detector CLIP-L, we see that incorporating the LLM already leads to a performance improvement ($+1.7\%$), likely due to the model's pretrained internal knowledge. However, adding the vision encoder leads to a further improvement ($+1.76\%$), as the LLM can now create a combined representation and rely on its vision encoder for general features, enhancing generalization performance (we use averages of 6 runs for the main stage 1 and 2 results to quantify random effects, $\sigma = 0.44\%$ here). In contrast, relying solely on the finetuned vision encoder yields significantly lower performance, as the model must learn specialized artifact representations itself, which quickly leads to overfitting. This shows that our dual encoders can improve upon both the binary SOTA detector and a finetuned LLM. We perform two control experiments: We use a simple pooling+concatenation approach for fusion. While this setup slightly outperforms the baseline on EFS, it overall performs worse than CLIP alone, likely due to overfitting. Secondly, we use a non-pretrained LLM for fusion (but trained vision encoder), which outperforms CLIP but performs worse than with pretraining ($92.63\%$). This shows that while most of the performance gain can be attributed to our dual-encoder fusion architecture, the LLM's pretrained knowledge also adds to the generalization ability.%

For stage 2, we find that exclusively relying on it is suboptimal because the weaker learning signal can not sufficiently incorporate both types of features, and relying on it at inference time also reduces performance due to degraded logit quality (see \cref{sec:inference_modes}).
In contrast, when starting from the strong base model of binary pretraining, RL finetuning can refine the learnt representation, improving performance to the best AUC of $93.50\% \pm 0.42\%$. Due to the small difference, we use averages of six runs here. The stage 2 models beat their respective paired stage 1 models in 5/6 cases, indicating a small but not yet statistically conclusive benefit. 

We therefore see that added capacity can quickly lead to overfitting if it is not avoided by providing frozen specialized features separately. Our dual-encoder architecture does that and significantly improves performance, with a small further improvement from using outcome-supervised rationale tuning, which also allows generating rationales if desired.
\begin{table}[tb]
    \small
    \centering
    \setlength{\tabcolsep}{3pt}
    
    \begin{tabular}{c|c:cc:cc|c} \toprule
     & LLM & DF-Enc. & V-Enc. & SFT & RL & avg. AUC \\ \hline
    Baseline & & \cmark & & \cmark & & 89.85\% \\ \hdashline
   \multirow{4}{*}{Ours Stage 1} & \cmark & & \cmark & \cmark & & 76.22\% \\
   & MLP & \cmark & \cmark & \cmark & & 87.80\% \\
     & \cmark & \cmark & & \cmark  & & 91.56\% \\
    & (\cmark) & \cmark & \cmark & \cmark & & 92.63\% \\
    & \cmark & \cmark & \cmark & \cmark & & 93.32\% \\ \hdashline
    \multirow{3}{*}{Ours Stage 2} & \cmark & \cmark & \cmark & & \cmark & 84.47\% \\
    & \cmark & \cmark & \cmark & \cmark & \cmark+inf. & 81.63\% \\
    & \cmark & \cmark & \cmark & \cmark & \cmark & \textbf{93.50\%} \\ \bottomrule
    \end{tabular}
    \caption{\textbf{Key component analysis.} Stage 1 shows the positive effect of our dual-encoder design. 
    Furthermore, using rationales as an additional training signal seems to have a small positive effect. However, when relying exclusively or during inference on RL, we observe reduced performance. This confirms our use of RL as auxiliary supervision only. ``+inf'' denotes using RL also at inference. (\cmark) refers to usage without pretraining.}
    \label{tab:abation-sft}
    \subfigurevspace
\end{table}

\begin{figure*}[tb]
    \centering
    \includegraphics[width=0.95\linewidth]{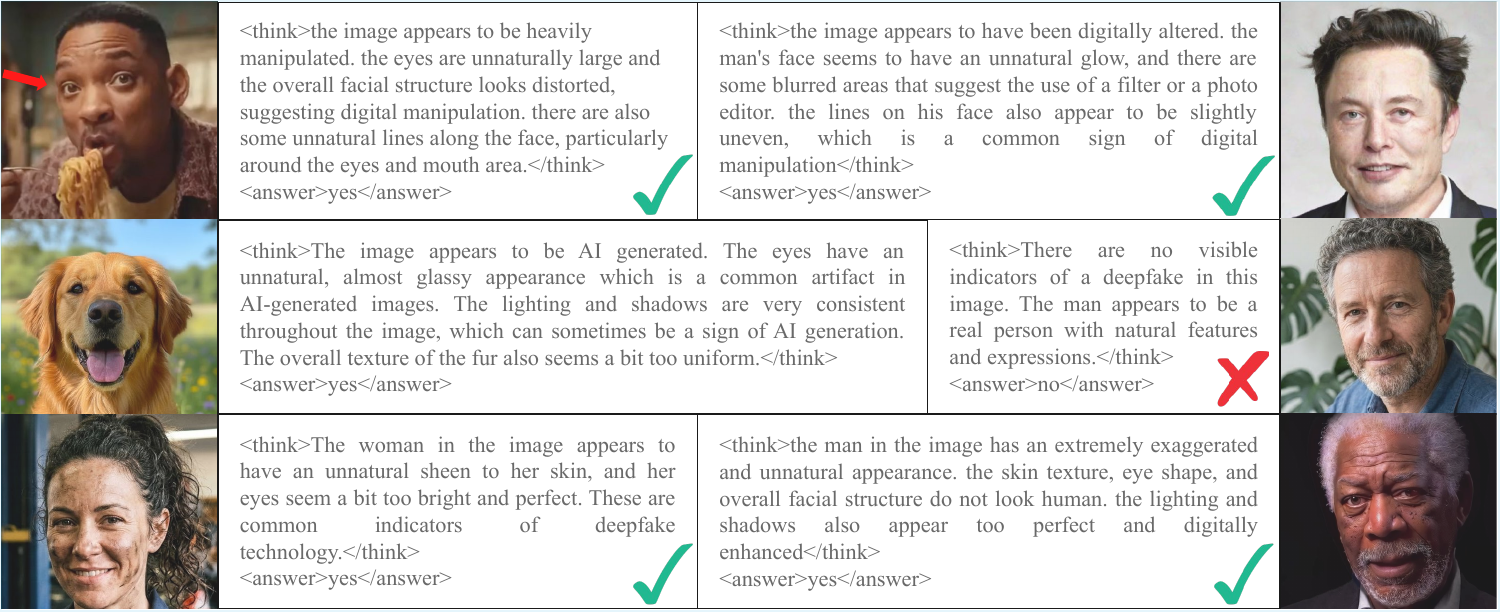}
    \caption{
    \textbf{Out-of-domain examples.} The first image is the famous Will-Smith-eating-spaghetti example~\cite{Wikipedia2025WillSmith} by Google's Veo 3~\cite{veo3}, the second one is taken from~\cite{Zou_2025NeuralPlagiarism}. The center left is generated by Google Gemini 2.5 Flash~\cite{GeminiNanoBanana}, the center right and lower left by Gemini 3 Flash~\cite{NanoBanana2}, and the final image is a still from~\cite{morganfreeman}. We manually added the red arrow to help the viewer locate the described artifact.
    }
    \label{fig:examples}
    \subfigurevspace
\end{figure*}

\vspace{2mm}
\noindent\textbf{Post-hoc reasoning.} We perform an experiment to isolate the effects of rationale generation compared to RL training. Since RL depends on randomness, we average 6 runs each. Our baseline SFT runs achieve an AUC of $93.32\%\pm 0.44\%$. We also run a control experiment, using RL but with inverted order (\ie the \texttt{<answer>} tag first, then \texttt{<think>}, details in the supplementary), such that causal masking prevents the rationale from changing the decision. Here we observe a reduction to $93.23\%\pm 0.47\%$, indicating that RL training by itself does not improve performance and may even be detrimental. We attribute this to some amount of forgetting \wrt the one-word answer evaluation protocol and no advantage to compensate for it. Similarly, continued SFT training at matched LoRA rank does not improve performance ($93.03\% \pm 0.64\%$), as the larger rank may overfit the training data. Rationale generation largely offsets the degradation caused by our answer-only policy-gradient setup and may provide a small additional benefit over SFT ($93.50\% \pm 0.42\%$).

\vspace{2mm}
\noindent\textbf{Shift of features due to RL.} To measure if RL actually shifts the model's internal features away from generator-specific low-level artifacts, we perform an experiment where we measure the shift in the distribution of features under the change of generator (change of low-level artifacts) and a shift of the underlying data (change of low and high-level features), by the Fréchet distance (using splits of DF40). We calculate the relative change for low-level features as $\frac{\Delta_{generator}}{\Delta_{data}}$. Here, our model gets a value of 0.525 before and 0.479 after RL finetuning. This proxy suggests reduced reliance on generator-specific low-level shifts after RL, consistent with the intuition that high-level features are necessary for description. The prompt is the same as for all evaluations.

\vspace{2mm}
\noindent\textbf{Causal grounding.}
To test image-grounding for the decision, we mask boxes around artifacts named in the explanation and compare to marking a matched face region, symmetric when possible.
On $N=50$ correctly classified examples, we measure the error $(1-p_y)$ for the true class. Masking the cited artifact region causes a $45\%$ larger error (geometric average) than masking the matched non-cited region, suggesting the explanations identify decision-relevant regions.%

\vspace{2mm}
\noindent\textbf{Out-of-Domain Examples.}
We analyze the explanations and decisions of our model on out-of-domain samples, using exemplary images from popular online deepfakes and state-of-the-art generators~\cite{veo3, Wikipedia2025WillSmith, morganfreeman, Zou_2025NeuralPlagiarism, GeminiNanoBanana, NanoBanana2} (\cref{fig:examples}). 
Our model can correctly identify the famous deepfake of Will Smith eating spaghetti based on an artifact at the edge of the face. This shows the descriptive quality of our method, despite the distribution shift from the training dataset. 
In the top right of \cref{fig:examples}, we show a neural plagiarism image. Our detector can recognize the image as fake and provide a reasonable explanation. This shows robustness to distribution shifts even beyond the realm of deepfakes. %
The image in the left center is a non-face image, so we use our model trained on the SID-Set~\cite{Huang2024SIDA}. This shows that our method can be applied to general images and will also learn descriptive explanations for non-face fakes.
In the right center, we show a failure case. This image is of particularly high quality and does not show any of the common diffusion artifacts (smooth textures, excessive symmetry, inconsistent lighting, ...). Therefore, our method fails to detect the image as a deepfake. Providing the image to ChatGPT gets the same result. This emphasizes the need for higher-quality datasets that encourage models to learn even more detailed signs.
In the lower left, we show an image of a person with oil on their face. Even under those conditions, our model can still identify the image as fake and provide a reasonable explanation. This shows its ability to generalize beyond the bounds of the training dataset.
Finally, on the lower right, we show a famous deepfake of Morgan Freeman. Although it is relatively old (2021), it remains a high-quality fake. Our model correctly identifies the deepfake, pointing to its ``exaggerated and unnatural appearance''. For a human observer, this would likely also be the ground for detection, as the image appears overly sharp and not like a natural photograph. More discussions, examples, and comparisons to ChatGPT are provided in the supplementary.

\vspace{2mm}
\noindent\textbf{Limitations. } While we show improved performance, neither our model's decisions nor rationales are perfect. Therefore, outputs should not be considered absolute truths, as false decisions can have negative effects. We do not test robustness to preprocessing, demographic shifts, \etc.

\section{Conclusion}
We presented a specialist–generalist deepfake detector in which an MLLM directly combines frozen forensic features with its native visual representations. This fusion accounts for the majority of the observed cross-domain improvement. We further introduced outcome-supervised rationale tuning, which learns free-form visual rationales from binary authenticity labels and provides a slight additional performance gain without requiring task-specific explanation annotations. Our model supports two deployment modes: direct binary prediction for standard detection and rationale generation when human-readable evidence is desired. Output-order and representation analyses suggest that rationale-before-answer training affects the learned decision function, although establishing the precise mechanism remains an open question. Overall, the results position visual rationales as optional auxiliary supervision for a strong multimodal detector rather than as a mandatory inference-time procedure.

\bibliography{main}

\clearpage
\setcounter{page}{1}
\setcounter{section}{0}
\setcounter{figure}{0}
\setcounter{table}{0}
\appendix

\renewcommand{\thetable}{\Alph{table}}
\renewcommand{\thefigure}{\Alph{figure}}
\renewcommand{\thesection}{\Alph{section}}

\twocolumn[{%
\renewcommand\twocolumn[1][]{#1}%
\centering
    {\huge
    \textbf{Specialist-Generalist Fusion with Outcome-Supervised Rationales \\ for Deepfake Detection}}
    \vspace{7mm}
    
    \large
    Supplementary Material
    \vspace{2mm}
\begin{center}
    \includegraphics[width=\linewidth]{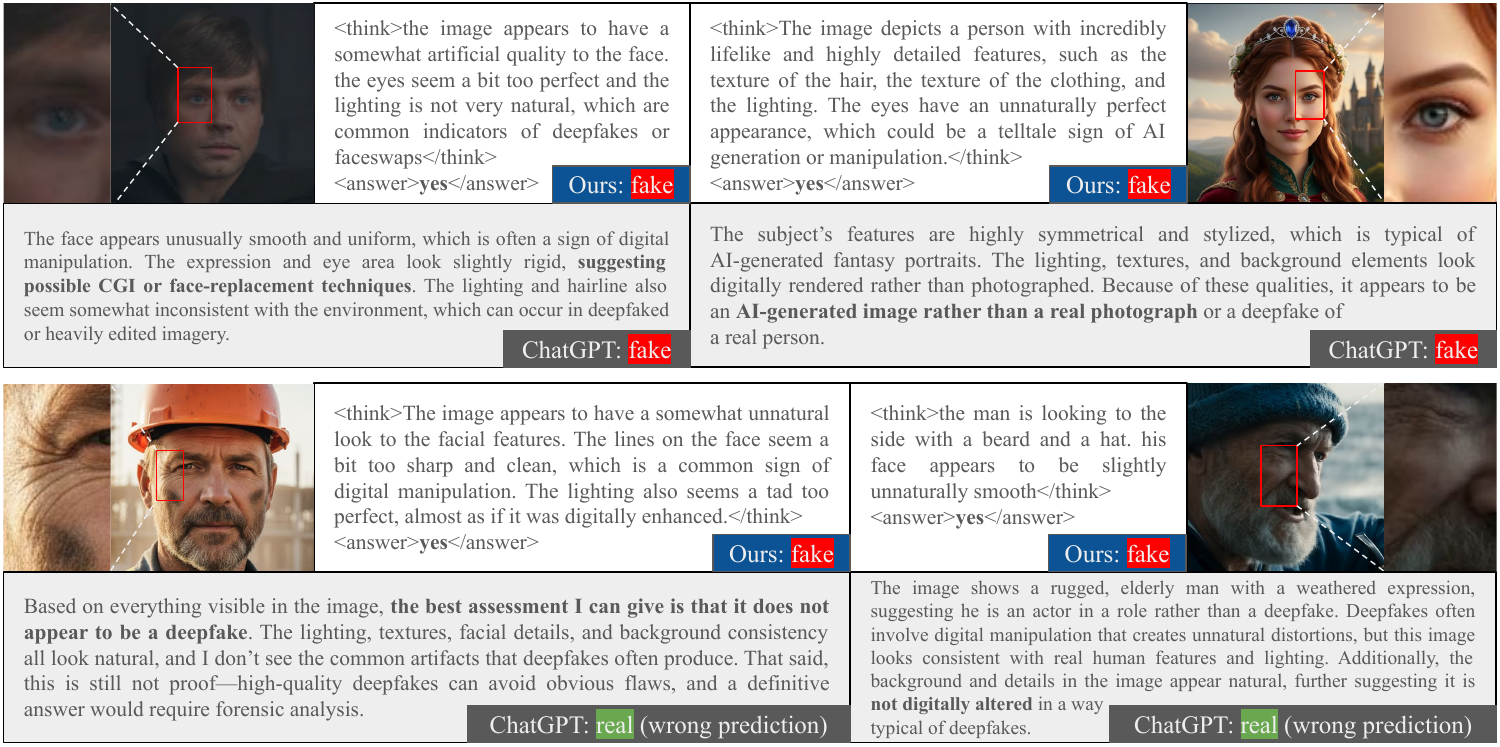}
    \captionof{figure}{\textbf{Samples from modern generators (all images are \colorbox{red}{\textcolor{white}{fake}}).} Even though artifacts are very slight, our model can detect and describe them. The first image is the famous Luke Skywalker Deepfake from The Mandalorian~\cite{Favreau_LucasMandalorian}, the second and third images are from~\cite{GeminiNanoBanana}. The final image is taken from the trailer video of Google's Veo 3~\cite{VeoSailor}. We also provide explanations by ChatGPT~\cite{OpenAI_2025} for comparison.}
    \label{fig:modern_suppl}
\end{center}
}]

\section*{Overview}

This document provides additional information that could not be included in the main paper. First, we provide more qualitative examples of our method on modern SOTA generators and compare to ChatGPT (\cref{sec:more_modern_generators}). Then, we give a full comparison to previous works on legacy datasets in~\cref{sec:legacy} and provide qualitative comparisons to other language-based deepfake detectors in~\cref{sec:sota_comparison}. Next, we provide details on the user study (\cref{sec:user_study_details}), computational requirements (\cref{sec:computational_requiremenst}), output control experiments (\cref{sec:reverse_reasoning}), and method parameters (\cref{sec:method_details}) as well as more qualitative samples on DF40~\cite{yan2024df40} (\cref{sec:additional_df40}) and compare our training/inference requirements to related work (\cref{sec:details_on_related_work}), as well as some theoretical background information (\cref{sec:background}). Finally, in~\cref{sec:on_the_definition_of_fake}, we provide some examples from related work on the usage of the term ``(deep)fake''.

\section{Samples from Modern Generators}
\label{sec:more_modern_generators}

\begin{figure*}
    \centering
    \includegraphics[width=0.6\linewidth]{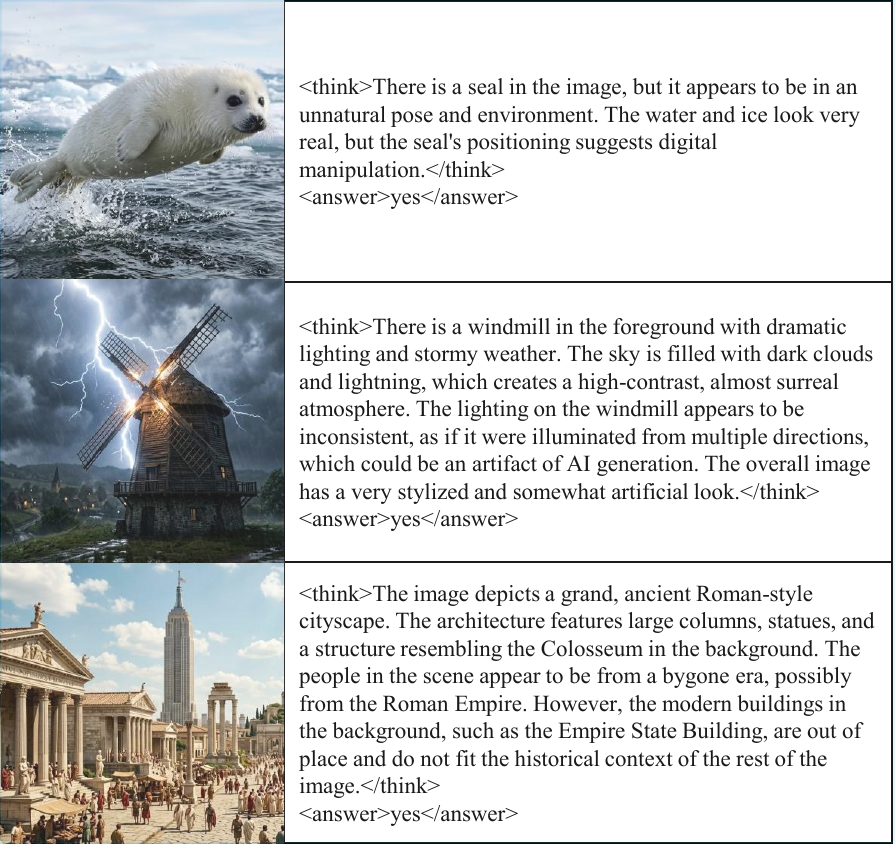}
    \caption{\textbf{AIGI samples from Gemini 3 Flash~\cite{NanoBanana2}}. All shown images are fake. Note how our model can integrate general real-world knowledge into its explanations.}
    \label{fig:AIGI-suppl}
\end{figure*}

We show some more examples from modern sources~\cite{Favreau_LucasMandalorian, GeminiNanoBanana, veo3} in~\cref{fig:modern_suppl}. Since the first image is a Hollywood-grade deepfake, it is hard for a human to detect that it is not real. It is therefore notable that our model manages to recognize it, based on some slight inconsistency in the lighting. We attribute this to the dual-encoder design using the specialist detector, allowing the detection of artifacts that are hard to spot -- even for humans.

The second and third images both show very high image quality, but they appear overly perfect, with all lines (eyelashes, wrinkles, ...) appearing not random enough for a real photograph. Our model can point out these details, indicating that the RL finetuning is capable of teaching the model to capture fine artifacts without supervision.

Finally, the last image appears to have a very slight blur on the man's cheek, which the model recognizes as a sign of a fake. We attribute this to the dual-encoder architecture and alignment: recognizing the small detail while being able to differentiate it from other types of blur, such as the blurry background, and then describing it using the general vision features. These images show the generalization ability of our model, even to very high-quality fakes. 

When comparing with explanations generated by ChatGPT~\cite{OpenAI_2025}, we observe that its explanations are relatively global (e.g., lighting, textures, ...), whereas our model can precisely pinpoint artifact areas. Furthermore, in two cases, ChatGPT does not detect the deepfake. This shows that for very high-quality deepfakes, finetuning is necessary, even when compared with very large commercial language models.

In \cref{fig:AIGI-suppl}, we show some examples of the generalization capabilities of our model trained on SID-Set~\cite{Huang2024SIDA}. The first image shows a seal jumping out of the water, which would not be uncommon for a dolphin, but is not something seals do, which our model recognizes. This shows that, despite being trained on SID-Set, which includes only one generator (so learning its specific artifacts is possible), our model still incorporates external knowledge, enabling it to also generalize to images from different generators. 

The second image shows lightning behind an old windmill. Still, the windmill is lit up from the front, which our model recognizes as a sign for AI generation. This exemplifies that our model can learn physical inconsistencies without direct supervision, underscoring the inherent capabilities of MLLMs due to their vast pretraining and multi-modal alignment.

Finally, in the last image, we test our model's general knowledge. We generate an image of ancient Rome, but include the Empire State Building in the background. Our model recognizes the fact that this combination should not exist and therefore flags the image as fake. We can, therefore, observe that our model prefers high-level semantic inconsistencies over low-level generator-specific ones.

These examples show that while our model was mainly developed for deepfakes of faces, it works equally well on general images. Furthermore, we see that it effectively learns to incorporate high-level artifacts and explain them correctly without the need for full supervision during training.

\section{Comparison on Legacy Datasets}
\label{sec:legacy}

For comparison with other language-based and binary deepfake detectors, we train on the FaceForensics++~\cite{roessler2019faceforensicspp} dataset and evaluate on CDFv2~\cite{Li2019CelebDFv2}, DFDC~\cite{Dolhansky2020DFDC}, and DFDCP~\cite{Dolhansky2019DFDCP}, following the established cross-dataset protocol~\cite{yu2025unlockingcapabilitieslargevisionlanguage, guo2025rethinkingvisionlanguagemodelface, Sun2023TowardsGV}. Since these datasets are no longer the current state-of-the-art, as they only include older methods and, \eg, no fully synthetic images, we only include the comparison because many detectors have been evaluated on these datasets. However, the comparison has limited practical implications, as it has been shown that performance on these datasets does not generalize well to more modern fakes~\cite{yan2024df40}. %

\Cref{tab:sota-ff-suppl} shows that our model outperforms all other language-based deepfake detectors on the most common CDFv2 dataset, as well as the particularly challenging DFDC dataset, while achieving second place on DFDCP. This indicates that while our model works particularly well on modern datasets, it still outperforms previous language-based deepfake detectors on old datasets, showing the flexibility of our technique. 

\begin{table*}[tb]
    \small
    \centering
    
    \begin{tabular}{c|c|ccc} \toprule
    & Model & CDFv2 & DFDCP & DFDC \\ \hline
    \multirow{15}{*}{Binary detectors} & Xception~\cite{roessler2019faceforensicspp, nguyen2024laa} & 61.18\% & 69.90\% & - \\ 
    & EfficientNet-B4~\cite{Tan2019EfficientNetRM, Yan2023DeepfakeBenchAC} & 74.87\% & 72.83\% & 69.55\% \\
    & FaceXRay+BI~\cite{Li2019FaceXRay} & 79.50\% & 80.92\% & 65.50\% \\
    & RECCE~\cite{Cao2022RECCE} & 68.71\% & - & 69.06\% \\
    & SPSL~\cite{Liu2021SPSL} & 76.88\% & 74.08\% & 70.40\% \\
    & UIA-ViT~\cite{Zhuang2022UIAViTUI} & 82.41\% & 75.80\% & - \\
    & SeeABLE~\cite{Larue2022SeeABLESD} & 78.30\% & 86.30\% & 75.90\% \\
    & UCF~\cite{Yan2023UCFUC} & 75.27\% & 75.94\% & 71.91\% \\
    & Controllable GS~\cite{Guo2023ControllableGF} & 84.97\% & 81.65\% & - \\
    & LSDA~\cite{Yan2023LSDA} & 83.00\% & 81.50\% & 73.60\% \\
    & SBI~\cite{Shiohara2022SBI} & 93.18\% & 86.15\% & 72.42\% \\
    & LAA~\cite{nguyen2024laa} & 95.40\% & 86.94\% & 71.70\% \\
    & PMM~\cite{hopf2025practicalmanipulationmodelrobust} & 91.53\% & \textbf{93.15}\% & 75.21\% \\
    & Effort~\cite{yan2025orthogonalsubspacedecompositiongeneralizable} & \underline{95.60\%} & 90.90\% & \textbf{84.20}\% \\
    & FakeSTormer~\cite{nguyen2025vulnerabilityawarespatiotemporallearninggeneralizable} & 92.40\% & 90.00\% & 74.6\% \\
    \hdashline
    \multirow{8}{*}{Language Detectors} & BLIP-TI-Xcep.~\cite{zhang2024commonsensereasoningdeepfake, roessler2019faceforensicspp} & 62.41\% & - & - \\
    & BLIP-TI-RECCE~\cite{zhang2024commonsensereasoningdeepfake, Cao2022RECCE} & 70.21\% & - & - \\  
    & BLIP-TI-HiFi~\cite{zhang2024commonsensereasoningdeepfake, guo2023hifi} & 71.00\% & - & - \\ 
    & BLIP-TI-SBI~\cite{zhang2024commonsensereasoningdeepfake, Shiohara2022SBI} & 93.98\% & - & - \\ 
    & VL-FFD~\cite{Sun2023TowardsGV} & 84.80\% & 84.74\% & - \\
    & UCLVLM~\cite{yu2025unlockingcapabilitieslargevisionlanguage} & 94.71\% & \underline{91.81\%} & 79.12\% \\
    & M2F2~\cite{guo2025rethinkingvisionlanguagemodelface} & 95.10\% & 87.80\% & - \\ \cline{2-5}
    & \textbf{Ours} & \textbf{95.75}\% & 90.29\% & \underline{82.75\%} \\ \bottomrule
    \end{tabular}
    \caption{\textbf{Comparison on legacy datasets.} Best values are highlighted in \textbf{bold}, second best values are \underline{underlined}. Our method performs comparably to previous work, even on these older datasets for which the compared methods were specifically designed.}
    \label{tab:sota-ff-suppl}
    \subfigurevspace
\end{table*}

Furthermore, we see that our method performs best on CDFv2~\cite{Li2019CelebDFv2}, which is the most commonly used benchmark dataset. On the other two datasets, it performs similarly to the top methods, indicating that our model obtains comparable performance despite not having been specifically tuned for these datasets. Also, it is known that DFDC~\cite{Dolhansky2020DFDC} and DFDCP~\cite{Dolhansky2019DFDCP} include low-quality images, and specifically designing a model for these low-quality conditions can be very helpful~\cite{hopf2025practicalmanipulationmodelrobust}. Since we aim for more modern datasets and explanations, we do not specifically tune our method for such circumstances and still achieve comparable performance. 

\section{Comparison to other Language Detectors}
\label{sec:sota_comparison}

\definecolor{ours}{RGB}{11,83,148}
\definecolor{vldfd}{RGB}{180,95,6}
\definecolor{uclvlm}{RGB}{125,29,63}
\definecolor{m2f2}{RGB}{166,28,0}
\definecolor{qwen}{RGB}{130,113,145}

\begin{figure*}[tb]
    \centering
    \includegraphics[width=\linewidth]{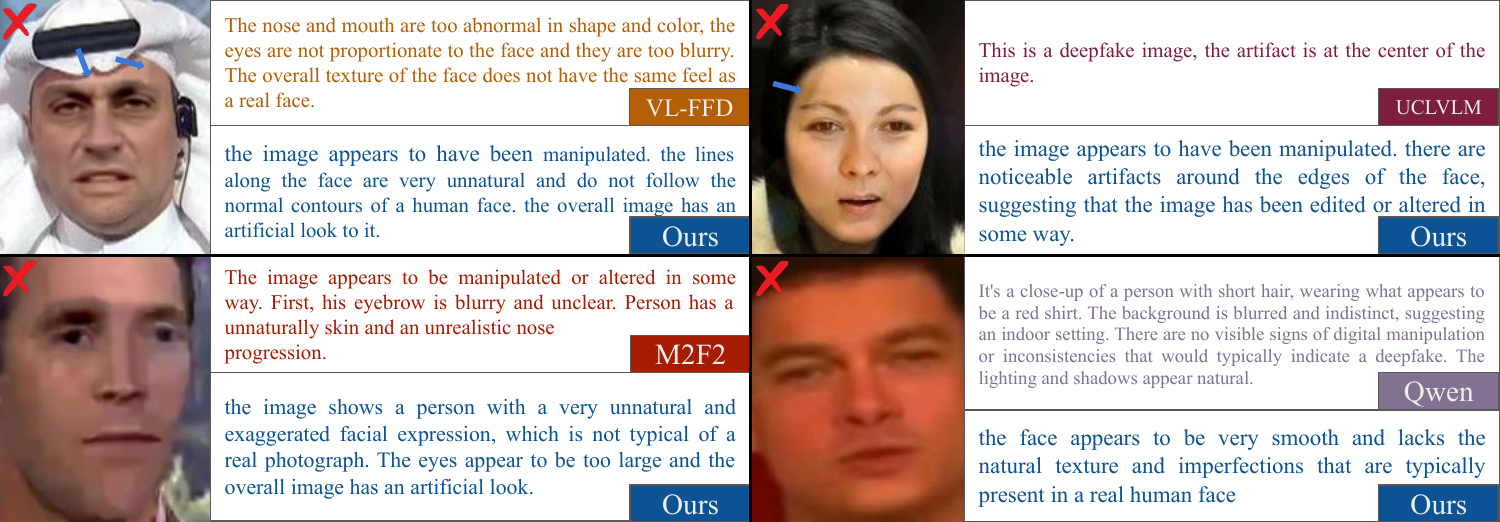}
    \caption{\textbf{Comparison of \colorbox{ours}{\textcolor{white}{our model}} to previous work} (\colorbox{vldfd}{\textcolor{white}{VL-FFD}}~\cite{Sun2023TowardsGV}, \colorbox{uclvlm}{\textcolor{white}{UCLVLM}}~\cite{yu2025unlockingcapabilitieslargevisionlanguage}, \colorbox{m2f2}{\textcolor{white}{M2F2}}~\cite{guo2025rethinkingvisionlanguagemodelface} and a baseline \colorbox{qwen}{\textcolor{white}{Qwen}}~\cite{bai2025qwen25vltechnicalreport}). We use the images and descriptions provided in the respective papers (except for Qwen) for a best-case scenario for the compared methods. Still, our model gives more useful answers, \eg providing information about ``unnatural lines'' (manually marked with arrows in the figure) where applicable. All shown images are fake.
    }
    \label{fig:comparison}
    \subfigurevspace
\end{figure*}

\begin{figure*}
    \centering
    \includegraphics[width=\linewidth]{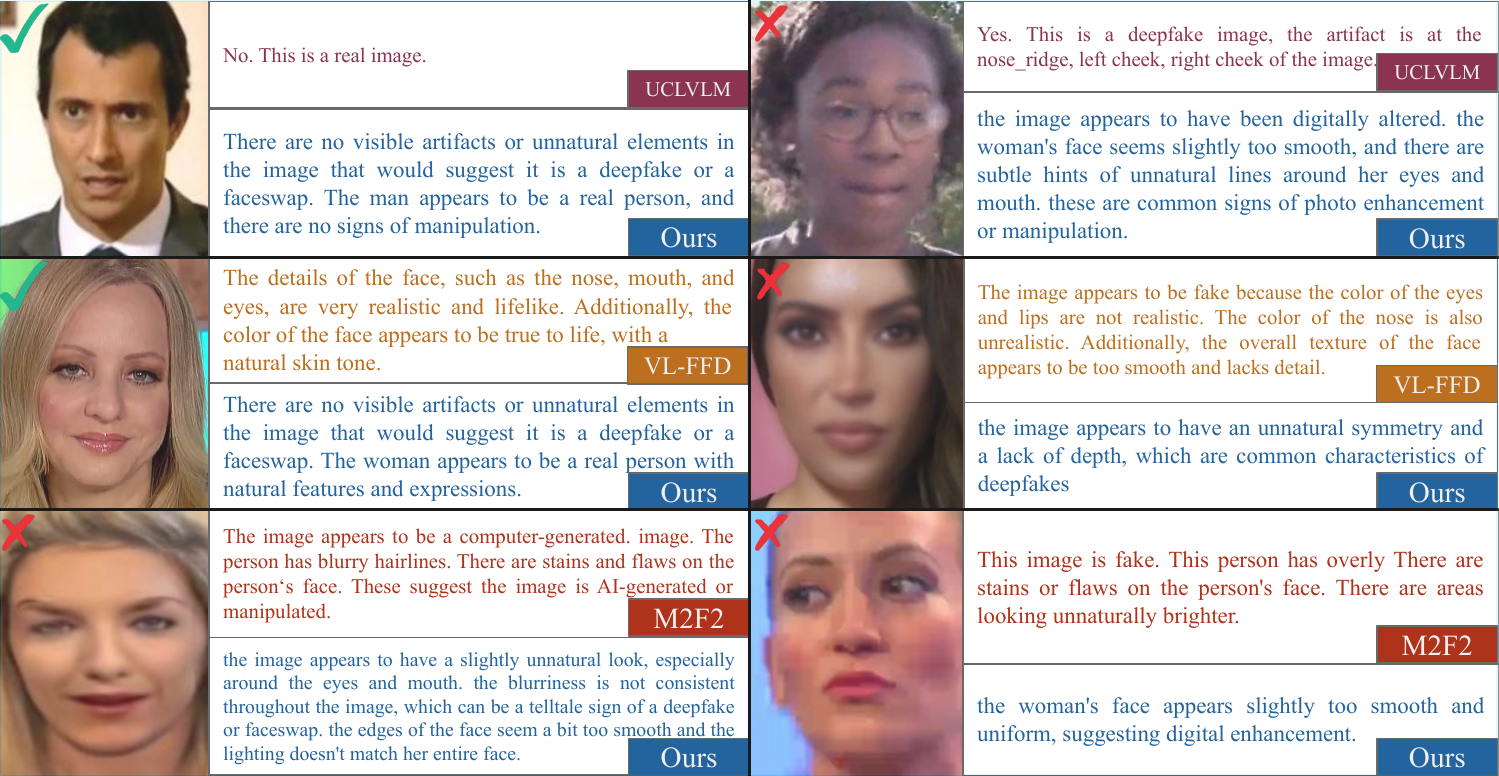}
    \caption{\textbf{More comparisons to SOTA language-based deepfake detectors.} We compare \colorbox{ours}{\textcolor{white}{our model}} to previous work (\colorbox{vldfd}{\textcolor{white}{VL-FFD}}~\cite{Sun2023TowardsGV}, \colorbox{uclvlm}{\textcolor{white}{UCLVLM}}~\cite{yu2025unlockingcapabilitieslargevisionlanguage}, \colorbox{m2f2}{\textcolor{white}{M2F2}}~\cite{guo2025rethinkingvisionlanguagemodelface}), using images directly from the respective papers. {\color{red}\xmark}:~deepfake, {\color{green}\cmark}:~real image.}
    \label{fig:comparison_suppl}
\end{figure*}

To compare to previous works~\cite{yu2025unlockingcapabilitieslargevisionlanguage, guo2025rethinkingvisionlanguagemodelface, bai2025qwen25vltechnicalreport, Sun2023TowardsGV} in terms of explanation quality, we show examples in \cref{fig:comparison} and \cref{fig:comparison_suppl}. For the deepfake detection papers, we take images directly from the respective paper, therefore using the cases that the authors considered best to showcase their models' performance. Despite being trained without full supervision, our explanations are similar or better, as our model can, for example, highlight blending boundaries, which are (if present) a reliable sign of a deepfake. The baseline Qwen~\cite{bai2025qwen25vltechnicalreport} model does not find meaningful features as it is trained for general image recognition, showing that finetuning is necessary. M2F2\cite{guo2025rethinkingvisionlanguagemodelface} provides comparable explanations to ours. However, they are limited to an annotated dataset and cannot easily be generalized to new data. VL-FFD\cite{Sun2023TowardsGV} gives descriptions, which, while linguistically high quality, are not clear signs of deepfakes (like a mouth being abnormal in shape), and some explanations also do not match the image contents. Finally, UCLVLM~\cite{yu2025unlockingcapabilitieslargevisionlanguage} only provides very short descriptions, which are limited to the location of the artifact, which is generally ``at the center of the image''. We therefore consider our explanations to be on par or better than the previous state-of-the-art, while only requiring binary labels for training instead of relying on supervision.

\section{User Study Details}
\label{sec:user_study_details}

For the user study, we asked three experts in the field of computer vision to judge the quality of our descriptions. Each participant was shown ten images and descriptions from DD-VQA~\cite{zhang2024commonsensereasoningdeepfake} and ten from our method, in random order. Users were asked to rate the descriptions on a scale from 1 (bad) to 5 (good). From these values, we calculate the mean and standard deviation. 

Given the small sample size, we verify the validity of our results using the bootstrap method~\cite{efron1979bootstrap}. We intend to show that our results are close to human labels, more specifically, that our mean opinion score is within one standard deviation of the reference mean opinion score.

We resample the results from the user study $1,000,000$ times with replacement. For our method and the reference (human-labeled) data, 
we then calculate the mean $\mu_{ours}, \mu_{ref}$ and standard deviation $\sigma_{ours}, \sigma_{ref}$. For the hardest test conditions, we set $\sigma = \min\{\sigma_{ours}, \sigma_{ref}\}$, so our score needs to be within the smaller of the two standard deviations. We then calculate the probability of $\mu_{ref} < \mu_{ours} + \sigma$. We find that probability to be $0.988$. So the number of samples is sufficient to support the claim that our results are reasonably close (\ie less than one $\sigma$) from human labels. We also test for $\mu_{ref} < \mu_{ours} + \sigma/2$ and find a probability of $0.767$, so we are likely even within half a standard deviation. On average, we find a distance of $0.34\cdot\sigma$ between the mean opinion scores.

\begin{figure*}
    \centering
    \includegraphics[width=\linewidth]{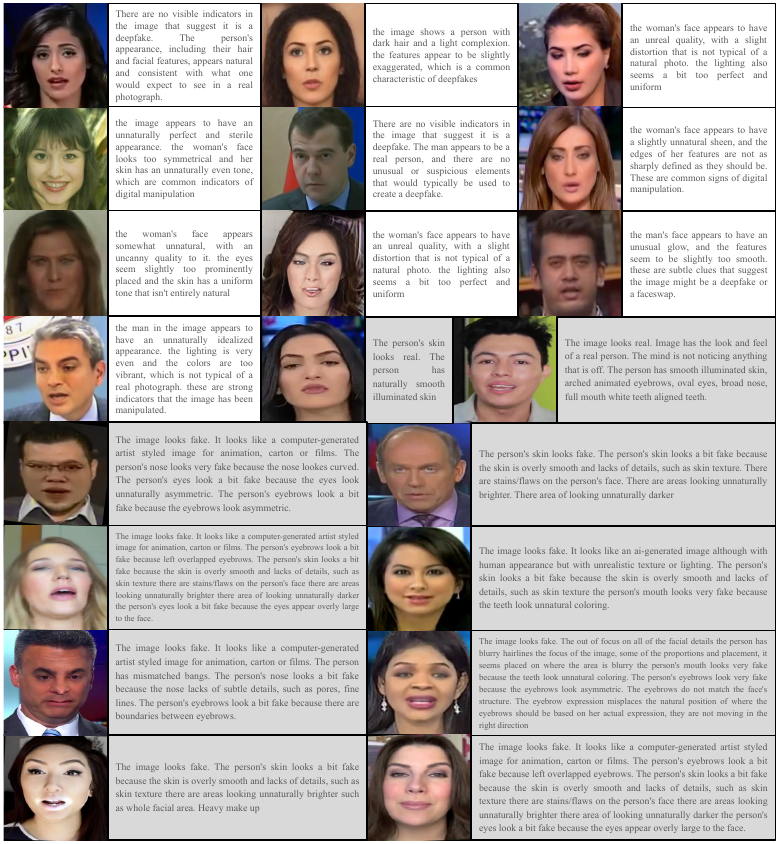}
    \caption{\textbf{Images and captions used in our user study.} The first ten captions (white) are from our method; the remaining ten (grey) are human annotations from the DD-VQA dataset~\cite{zhang2024commonsensereasoningdeepfake}. All final decisions are correct, in order to only compare explanation quality.}
    \label{fig:user_study}
\end{figure*}

This shows that the quality of our explanations is roughly comparable to human annotations.
The full set of images used is shown in~\cref{fig:user_study}, to allow the reader to judge for themselves. We selected images for which the classification decision was taken correctly, as we only intend to measure the explanation quality of our method and not its detection performance, for which we have strong quantitative experiments.

\section{Computational Requirements}
\label{sec:computational_requiremenst}

Given that our model utilizes a large language model, it requires significantly more computational resources compared to more lightweight binary detectors. The computational cost for different configurations of our model is shown in \cref{tab:compute}.
\begin{table}[tb]
    \centering
    
        \begin{tabular}{cc|ccc} \toprule
            Mode & training & Param. & trainable & TFLOPs \\ \midrule
            SFT & \cmark & 8.6B & 17M & 16.8 \\
            SFT & \xmark & 8.6B & - & 5.6 \\
            RL & \cmark & 8.7B & 107M & 178 \\ 
            RL & \xmark & 8.7B & - & 26.2\\
            \bottomrule
        \end{tabular}
        \caption{\textbf{Computational requirements.} RL training requires significantly more FLOPs because of multiple autoregressive generations per input image.}
    \label{tab:compute}
\end{table}
The difference in parameters between the SFT and RL versions stems from the use of different-sized LoRA adapters (ranks 4 and 32, respectively). We also see that the number of FLOPs during inference in RL mode (\ie when generating descriptions) is significantly lower than during training. This happens because during training, we need to sample multiple sequences to have a group of answers to compare to. We use $G = 4$ samples per group. Since the answer length is stochastic, the number of FLOPs in RL mode is not fixed. The shown values are averages of 30 groups. 

During training, backward passes and parameter updates have to be calculated, which is already included in the numbers.

\vspace{2mm}
\noindent\textbf{Hardware.} We train on a Linux-based SLURM cluster using NVIDIA L40 and L40s GPUs. We use 8 CPU cores and 64GB of memory, as well as one or three GPUs for stages 1 and 2, respectively. We implement our algorithm using \texttt{pytorch} and \texttt{transformers} in the framework of DF40~\cite{yan2024df40}.

\section{Stage 2 Control Experiments}
\label{sec:reverse_reasoning}

Here we detail the control experiments performed to separate the effects of rationale-generation from the change in objective (next-token-prediction to RL).

Our first experiment consists of reversing the output order. That is, we use the same experimental settings as before, but the requested output format is now reversed (\ie \begin{verbatim}
    <answer>[a-z]+</answer>
    <think>.*</think>
\end{verbatim} prompts are phrased accordingly). While this setup still uses RL, it does not enforce the usage of describable artifacts due to causal masking. By the time artifact descriptions are needed, the decision is already taken. Since only the decision is supervised, the explanation is free to hallucinate. When testing, we measure a performance of $93.23\% \pm 0.47\%$, which is even lower than the SFT-only result of $ 93.32\% \pm 0.44\%$ (averages and standard deviations of 6 runs, since differences are small). With the correct order, however, the RL stage improves to an average of $93.50\% \pm 0.42\%$, beating its baselines in 5/6 cases. This indicates that RL by itself is not advantageous and may, in fact, be detrimental, which we attribute to some degree of forgetting of the one-word-answer output protocol, as the model is now trained with a different output format. Note that, as with all quantitative experiments, no reasoning was generated during testing, so performance differences are restricted to feature selection, as the answer pattern was not used during testing. We also perform an experiment with no reasoning (\ie standard stage 2 structure, but the \texttt{<think>} tags need to be empty). Here we see similarly reduced performance ($93.28\% \pm 0.51\%$), further strengthening the claim that generating free-form rationales is responsible for improvements and not RL itself. 

\section{Method Details}
\label{sec:method_details}

For each input image, we generate a group of $G = 4$ sample answers using a temperature of $\tau = 1$. Limited by our GPUs' 48GB of VRAM, we generate a maximum answer length of 120 new tokens. The model learns to generate fewer than the set number of new tokens, as running into the generation limit will prevent it from completing the format requirements (\ie finishing with \texttt{</answer>}). Furthermore, we randomly select one out of five possible prompts for each input image in order to avoid overfitting to a specific prompt. We leave the curation of a dataset of potential prompts for further improved diversity to future work.

During testing, we align with the common protocols~\cite{Shiohara2022SBI, nguyen2024laa, yan2024df40} of providing video-level statistics for the legacy datasets and frame-level statistics for DF40.

\Cref{tab:additional_aucs} provides more detailed AUCs for some baseline configs from the ablation. We observe that rationale-tuning consistently improves performance on FR while slightly reducing performance on FS.

\begin{table}[tb]
    \centering
    \begin{tabular}{c|ccc}\toprule
        Model & FS$\uparrow$ & FR$\uparrow$ & EFS$\uparrow$ \\ \midrule
        
        Qwen & 68.17\% & 83.92\% & 76.98\% \\
        Reasoning inference & 80.84\% & 79.37\% & 84.69\%\\
        SFT / binary alignment & \textbf{93.92\%} & 90.28\% & 95.76\% \\
        
       \textbf{Ours} & 93.38\% & \textbf{91.30\%} & \textbf{95.83\%}\\
        \bottomrule
    \end{tabular}
    
    \caption{\textbf{Baselines} on DF40 splits, in line with \cref{tab:sota}. These are also covered in the ablation, but we provide additional detailed AUCs on these relevant baseline configurations.}
    \label{tab:additional_aucs}
    \subfigurevspace
\end{table}

\subsection{Meaning of $\alpha$}

During training, we use $\alpha = 0.01$ in order for the format reward to dominate correctness, which aids stability in the early stages of training. Here we prove that once the format is properly followed (\ie $r^f_i = r^f_j \ \forall i, j \in \mathcal{G}$), the choice of $\alpha$ becomes irrelevant.

\begin{proof}
    \allowdisplaybreaks
    Let $r^f_i = r^f_j \ \forall i, j \in \mathcal{G}$. Then $A_g$ is given as follows:
    \begin{align}
        A_g &= \frac{r_g - \text{mean}_g(r)}{\text{std}_g(r)} = \frac{r_g - \frac{1}{G}\sum_{j=1}^G r_j}{\sqrt{\frac{1}{G-1} \sum_{j=1}^G \left( r_j - \frac{1}{G} \sum_{i=1}^G r_i \right)^2}} \\
        &= \frac{r_g^f + \alpha r_g^c - \frac{1}{G}\sum_{j=1}^G \left( r_j^f + \alpha r_j^c \right)}{\sqrt{\frac{1}{G-1} \sum_{j=1}^G \left( r_j^f + \alpha r_j^c - \frac{1}{G} \sum_{i=1}^G \left( r_i^f + \alpha r_i^c \right) \right)^2}} \\
        &= \frac{r^f + \alpha r_g^c - \frac{1}{G}\sum_{j=1}^G \left( r^f + \alpha r_j^c \right)}{\sqrt{\frac{1}{G-1} \sum_{j=1}^G \left( r^f + \alpha r_j^c - \frac{1}{G} \sum_{i=1}^G \left( r^f + \alpha r_i^c \right) \right)^2}} \\
        &= \frac{r^f + \alpha r_g^c - \frac{1}{G}\sum_{j=1}^G r^f - \frac{1}{G}\sum_{j=1}^G \alpha r_j^c}{\sqrt{\frac{1}{G-1} \sum_{j=1}^G \left( r^f + \alpha r_j^c - \frac{1}{G} \sum_{i=1}^G r^f - \frac{1}{G}\sum_{i=1}^G \alpha r_i^c \right)^2}} \\
        &= \frac{r^f + \alpha r_g^c - r^f - \alpha\frac{1}{G}\sum_{j=1}^G r_j^c}{\sqrt{\frac{1}{G-1} \sum_{j=1}^G \left( r^f + \alpha r_j^c - r^f - \alpha\frac{1}{G}\sum_{i=1}^G r_i^c \right)^2}} \\
        &= \frac{\alpha \left( r_g^c - \frac{1}{G}\sum_{j=1}^G r_j^c \right)}{\sqrt{\frac{1}{G-1} \sum_{j=1}^G \left( \alpha \left(r_j^c - \frac{1}{G}\sum_{i=1}^G r_i^c \right)\right)^2}} \\
        &= \frac{\alpha \left( r_g^c - \frac{1}{G}\sum_{j=1}^G r_j^c \right)}{\sqrt{\frac{1}{G-1} \sum_{j=1}^G \alpha^2 \left( r_j^c - \frac{1}{G}\sum_{i=1}^G r_i^c \right)^2}} \\
        &= \frac{\alpha \left( r_g^c - \frac{1}{G}\sum_{j=1}^G r_j^c \right)}{\sqrt{ \alpha^2 \frac{1}{G-1} \sum_{j=1}^G \left( r_j^c - \frac{1}{G}\sum_{i=1}^G r_i^c \right)^2}} \\
        &= \frac{\alpha \left( r_g^c - \frac{1}{G}\sum_{j=1}^G r_j^c \right)}{\alpha \sqrt{ \frac{1}{G-1} \sum_{j=1}^G \left( r_j^c - \frac{1}{G}\sum_{i=1}^G r_i^c \right)^2}} \\
        &= \frac{r_g^c - \frac{1}{G}\sum_{j=1}^G r_j^c}{\sqrt{ \frac{1}{G-1} \sum_{j=1}^G \left( r_j^c - \frac{1}{G}\sum_{i=1}^G r_i^c \right)^2}} \\
        &= \frac{r_g^c - \text{mean}_g(r^c)}{\text{std}_g(r^c)}
    \end{align}

    Which does not depend on $\alpha$ anymore.
\end{proof}

\section{Additional DF40 Samples}
\label{sec:additional_df40}

\begin{figure*}[tb]
    \centering
    \includegraphics[width=\linewidth]{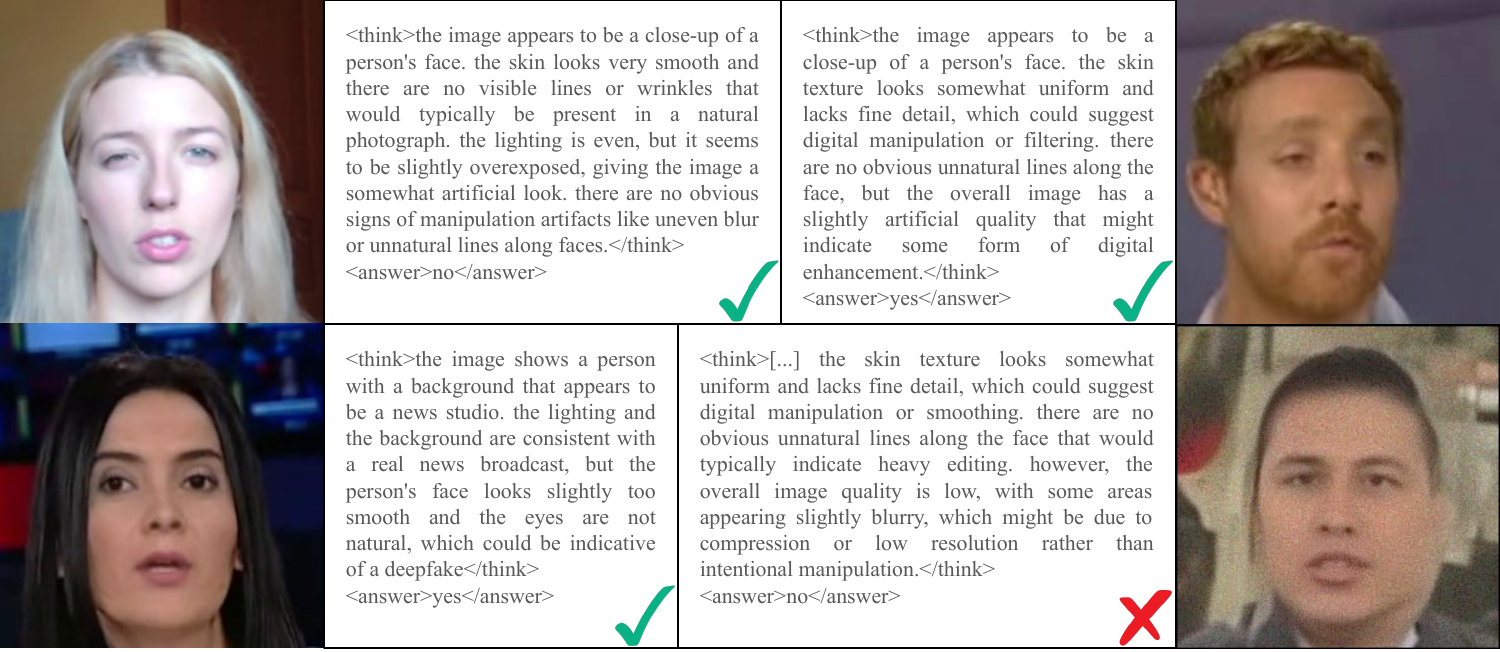}
    \caption{\textbf{Sample explanations based on DF40 images.} Note that even in the case of a wrong final decision, the explanation can still be helpful to the user. {\color{red}\xmark}:~wrong decision, {\color{green}\cmark}:~correct decision.}
    \label{fig:df40_examples}
\end{figure*}

In \cref{fig:df40_examples}, we show additional sample explanations based on the DF40~\cite{yan2024df40} dataset. We include a failure case to demonstrate that even an incorrect detection can provide meaningful insights for the user. In that case, the model notices uniform skin texture, which can be a sign of a fake, but given the lack of blending boundaries \etc it finally decides ``real''. Such an output still provides the user with helpful information, allowing them to decide for themselves if these signs are sufficient for considering the image fake. Note that our model provides arguments for both sides in multiple cases. In addition to enhancing interpretability, this also shows that our model does not simply sample explanations that correlate with the answer, but actually examines the image.

\section{Training and inference requirements of related work}
\label{sec:details_on_related_work}

\Cref{tab:related_work} shows a comparison of our method to related work in terms of its requirements for training and inference. Our model and RAIDX are able to work without ground-truth textual annotations. However, RAIDX requires retrieval and only works on fully-synthetic images, making it significantly less flexible. Not included in the table are two methods that do not fit the direct LLM framework: ReAlign~\cite{huang2026realigngeneralizableimageforgery}, which uses an RL-trained LLM as a teacher for a smaller, binary-only model, and EvoGuard~\cite{zhu2026evoguardextensibleagenticrlbased}, which builds an agentic setup to orchestrate different deepfake detection tools. 

\begin{table*}[tb]
    \centering
    
    \begin{tabular}{c|c:c:c:c} \toprule
        Name & Text Annotations & Faketype & Post-hoc & @inference \\ \midrule
        VL-FFD~\cite{Sun2023TowardsGV} &  \xmark & Faces & \xmark & n/a \\
        UCLVLM~\cite{yu2025unlockingcapabilitieslargevisionlanguage} & \xmark & Faces & \xmark & n/a  \\
        M2F2~\cite{guo2025rethinkingvisionlanguagemodelface} &  \xmark & Faces & \xmark & n/a \\
        AuthGuard~\cite{shen2025authguardgeneralizabledeepfakedetection} & \xmark & Faces & \xmark & n/a \\
        PRPO~\cite{nguyen2026prpoparagraphlevelpolicyoptimization} & \xmark & Faces & - & \xmark \\
        MARE~\cite{xu2026maremultimodalalignmentreinforcement} & \xmark & Faces & - & \xmark \\
        \hdashline
        RAIDX~\cite{Li2025RAIDXAR} & - & fully synth. & - & \xmark \\
        Veritas~\cite{tan2025veritasgeneralizabledeepfakedetection} & \xmark & fully synth. & - & \xmark \\
        ThinkFake~\cite{huang2025thinkfakereasoningmultimodallarge} & \xmark & fully-synth. & - & (\xmark) \\
        So-Fake~\cite{huang2025sofakebenchmarkingexplainingsocial} & \xmark & All & - & \xmark \\
        Ivy-Fake~\cite{jiang2026ivyfakeunifiedexplainableframework} & \xmark & All & - & \xmark \\
        Omni-Fake~\cite{Li_2026_OmniFake} & \xmark & All & - & \xmark \\
        REFORM~\cite{zhang2026cultivatingforensicreasoninggeneralizable} & \xmark & All & separate & n/a \\
        \midrule
        \textbf{Ours} & - & All & - & - \\ \bottomrule
    \end{tabular}
    \caption{\textbf{Training and inference requirements of our model and related work.} \textit{Text Annotations} refers to a supervised stage during training, regardless of whether the supervision comes from dataset annotation or some heuristic. \textit{@inference} refers to using reasoning at inference, rather than only for regularization. (\xmark) refers to using reasoning on some samples and not on others; n/a refers to a separate detection head, so reasoning is independent from detection.}
    \label{tab:related_work}
\end{table*}

\section{Background}
\label{sec:background}

This section provides a brief overview of techniques we use to supervise free-form rationales without ground-truth descriptions.

\subsection{Causal Attention} %

To avoid hallucinations, we need to generate the answer after the description due to the causal structure of language output. In causal language modelling, earlier tokens influence later ones, but not the other way around~\cite{Vaswani2017AttentionIA}. More formally: \begin{equation}
\label{eq:causallm}
    p(o | x, t) = \prod_{n=1}^N p(o_i|x, t, o_{<i})
\end{equation}
for input tokens $t$, image $x$ and output $o$ of length $N$. Therefore, if the decision is provided first, the explanation is conditioned on the decision, which can lead to hallucination. Therefore, the explanation needs to be provided \textit{first} and the decision needs to be taken \textit{afterwards}.

\subsection{RL for rationale training}
We start from group-relative policy optimization~\cite{shao2024deepseekmathpushinglimitsmathematical} as the reinforcement learning algorithm of our choice, as it is comparably lightweight. We sample a group of $G$ possible answers and calculate a reward $r_g, \ g\in \{1,..., G\}\eqqcolon \mathcal{G}$ for each answer. 
Given a group of rewards, we normalize to obtain the advantage $A_g$ as in \cite{deepseekai2025deepseekr1incentivizingreasoningcapability}:  
\begin{equation}
    A_g = \frac{r_g - \text{mean}_g(r)}{\text{std}_g(r)}
\end{equation}
setting $A_g = 0 \ \forall g \in \mathcal{G} $ if $r_i = r_j \ \forall i, j \in \mathcal{G}$.

To save on VRAM, we modify the algorithm slightly, setting $\pi_{\theta_{old}}(\cdot) = \text{sg}(\pi_\theta(\cdot))$, where $\text{sg}$ is the stop-gradient operator, avoiding the need to keep an old model in memory. This is a simplified case of the original algorithm, which includes updating the policy model. Our version is equivalent to a full update at every step, with the added advantage of reducing VRAM consumption. The main loss term then becomes 
\begin{equation}
\begin{split}
        \mathcal{L_\mathit{\mathit{RL}}} = -\frac{1}{G}\sum_{g=1}^{G} &\min\left(  \frac{\pi_\theta(o_g | q)}{\text{sg}(\pi_\theta(o_g | q))} \cdot A_g, \right. \\
        & \quad \left. \text{clip}\left( \frac{\pi_\theta(o_g | q)}{\text{sg}(\pi_\theta(o_g | q))} , 1\pm\varepsilon \right) \cdot A_g \right)
\end{split}
\end{equation} 
for group size $G$, and clipping range $\varepsilon=0.2$ following~\cite{deepseekai2025deepseekr1incentivizingreasoningcapability}. In this case, the clipping term is never active, since the old and new policies are the same, and could therefore be removed. 

Therefore, answers with larger rewards are encouraged, and ones with smaller rewards (\ie negative advantages) are discouraged. 
The final loss is then given by \begin{equation}
    \mathcal{L} = \mathcal{L}_\mathit{\mathit{\mathit{RL}}} + \beta \cdot \mathcal{L}_\mathit{reg}
\end{equation}
where $\beta$ is a hyperparameter, and $\mathcal{L}_{reg}$ is the KL-divergence based regularization term as in~\cite{shao2024deepseekmathpushinglimitsmathematical}.

\section{On the Definition of Fake}
\label{sec:on_the_definition_of_fake}

\begin{figure}
    \centering
    \includegraphics[width=\linewidth]{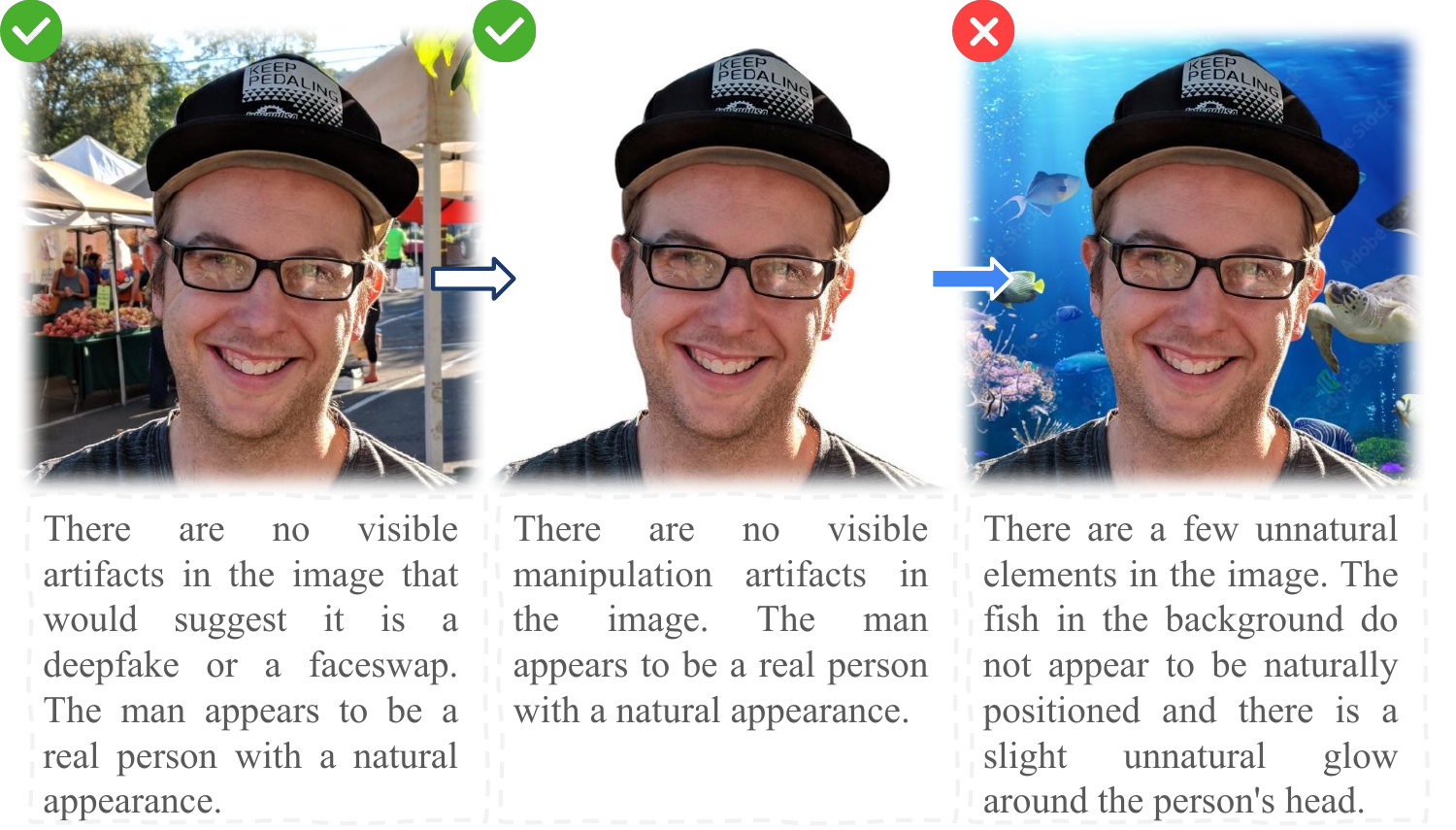}
    \caption{\textbf{What is a fake?} There exists a grey area between fakes and real images, \eg in the case of background removal. Our exemplary testing indicates that our model, in this case, bases its decision on the practical realism of the content of the image. We only print our model's descriptions, omitting format and decision.}
    \label{fig:background}
    \vspace{-2mm}
\end{figure}

The definition of deepfake remains fundamentally ambiguous. The prevailing view frames an image as fake primarily when it is AI-generated. Under this perspective, an untouched photograph is unquestionably real, while face swaps or fully synthetic images are fake. Yet such a dichotomy collapses under closer scrutiny. For example, is a background change, routine in videoconferencing, considered fake? And if so, by what standard?

Our observations underscore this ambiguity. Using an image from~\cite{Wadhwa_2018FaceImages} (see \cref{fig:background}), we compare background removal and replacement. A plain white background is often judged as real by most human observers, and our proposed model, which also joins the conclusion from commercialized GPT. Yet this perception of ``realness” is semantically questionable: the removal may be AI-assisted, and lighting inconsistencies undermine physical plausibility, \ie the lighting on the face suggests that a pure white background is not possible. By contrast, placing the same person underwater immediately feels fake -- not because of the editing tool, but because the scene could not have occurred.

These examples expose a deeper tension:
Should fakes be defined by the process (AI-generated) or the outcome (depicting events that never occurred)? Current definitions conflate these criteria, resulting in unstable and context-dependent judgments. This highlights the need to clarify and standardize what we mean by ``fake” in the first place.

Here we compile some examples from related work to show that the definition of ``fake'' is by no means as clear as it may seem. While the term ``deepfake'' might indicate the usage of some ``deep model'', this does not always need to be the case. The main training dataset FaceForensics++~\cite{roessler2019faceforensicspp} includes FaceSwaps~\cite{faceswap}, which do not use deep generative methods. Furthermore, some works~\cite{Li2019FaceXRay, Shiohara2022SBI} create pseudo-fakes, which directly blend two faces, without an intermediate generative step. While these are \textit{pseudo}-fakes, they are used to train detectors, indicating that they should be considered fake. Specifically~\cite{Shiohara2022SBI} introduces inconsistencies in sharpness and color between face and background, indicating that local manipulations are considered fake. Several works~\cite{nguyen2024laa, hopf2025practicalmanipulationmodelrobust, Yan2023DeepfakeBenchAC} consider such manipulations on the full image to not constitute a fake. Even changes made by deep models are not necessarily considered fake, as \eg superresolution is not considered deepfake in~\cite{yan2024df40}, but only described as creating similar generative artifacts. 

Finally, for the case of background removal, the FaceForensics++ dataset~\cite{roessler2019faceforensicspp} includes several ``real'' videos, taken from TV broadcasts with replaced backgrounds (probably using a green screen). 

This overview is not meant to give a final answer to what should be fake, but to spark a discussion, which may lead to a standardized definition of the term ``(deep)fake''.

\end{document}